\documentclass[lettersize,journal]{IEEEtran}
\usepackage{amsmath,amsfonts}
\usepackage{algorithm}
\usepackage{algpseudocode}
\usepackage{array}
\usepackage[caption=false,font=scriptsize,labelfont=sf,textfont=sf]{subfig}
\usepackage{textcomp}
\usepackage{stfloats}
\usepackage{url}
\usepackage{verbatim}
\usepackage{graphicx}
\usepackage{cite}

\usepackage{multirow}
\usepackage{pifont}
\usepackage{booktabs}
\usepackage[dvipsnames,svgnames,x11names]{xcolor}
\usepackage{makecell}
\newcommand{\sz}[1]{\textcolor{black}{#1}}%red
\graphicspath{{img/}}

\algrenewcommand\algorithmicprocedure{\textcolor{black}{\textbf{procedure}}}
\algrenewtext{For}[1]{\textbf{\textcolor{black}{for #1 do}}}
\algrenewtext{EndFor}{\textbf{\textcolor{black}{end for}}}

\begin{document}

\title{A Unified Hierarchical Framework for Fine-grained Cross-view Geo-localization over Large-scale Scenarios}

\author{Zhuo Song, Ye Zhang, Kunhong Li, Longguang Wang, Yulan Guo,~\IEEEmembership{Senior Member, IEEE}
\thanks{Z. Song, Y. Zhang, K. Li, and Y. Guo are with the School of Electronics and Communication Engineering, the Shenzhen Campus of Sun Yat-sen University, Sun Yat-sen University, Shenzhen, China, 518107. (E-mail: songzh28@mail2.sysu.edu.cn; zhangy2658@mail.sysu.edu.cn; likh25@mail2.sysu.edu.cn; guoyulan@sysu.edu.cn).}
\thanks{Longguang Wang is with the College of Electronic Science and Technology, Aviation University of Air Force, Changchun, China, 130012. (E-mail: wanglongguang15@nudt.edu.cn).}% <-this % stops a space
\thanks{Corresponding authors: Ye Zhang, Yulan Guo.}}

\markboth{Journal of \LaTeX\ Class Files,~Vol.~14, No.~8, August~2021}%
{Song \MakeLowercase{\textit{et al.}}: A Unified Hierarchical Framework for
Fine-grained Cross-view Geo-localization over Large-scale Scenarios}

\maketitle

\begin{abstract}
Cross-view geo-localization is a promising solution for large-scale localization problems, requiring the sequential execution of retrieval and metric localization tasks to achieve fine-grained predictions.
However, existing methods typically focus on designing standalone models for these two tasks, resulting in inefficient collaboration and increased training overhead.
In this paper, we propose UnifyGeo, a novel unified hierarchical geo-localization framework that integrates retrieval and metric localization tasks into a single network.
Specifically, we first employ a unified learning strategy to jointly learn multi-granularity representations,  
\sz{establishing task associations between retrieval and metric localization.}
Subsequently, we design a re-ranking mechanism guided by a dedicated loss function, which enhances geo-localization performance by improving both retrieval accuracy and metric localization references.
Extensive experiments demonstrate that UnifyGeo significantly outperforms state-of-the-art methods in both task-isolated and task-associated settings.
\sz{On the challenging VIGOR benchmark, UnifyGeo achieves 39.64\% and 25.58\% 1-meter-level localization recall under same-area and cross-area evaluations, respectively, demonstrating strong fine-grained localization capability in large-scale scenarios.}
\textcolor{black}{Code will be available at https://github.com/chord-sz/UnifyGeo.}
\end{abstract}


\section{Introduction}
\IEEEPARstart{C}{ross}-View Geo-Localization (CVGL) aims to determine the geographic location of a ground-level image by matching it to a database of geo-tagged aerial or satellite imagery \cite{shi2019spatial, song2024unified, 10797651, 10376356}.
The widespread availability and global coverage of satellite imagery make CVGL a promising approach for scalable, large-scale geo-localization, which is crucial for various applications, including autonomous driving \cite{doan2019scalable, kim2017satellite}, street navigation \cite{mirowski2018learning, shetty2019uav}, 
augmented reality \cite{chiu2018augmented}, and urban computing \cite{zou2025deep}.
However, the growing demand for real-world applications has placed increasingly stringent requirements on positioning accuracy.
To achieve such fine-grained geo-localization, it is essential to address the significant appearance and geometric disparities between ground and aerial views, while overcoming the
inherent visual ambiguities across large geographical areas \cite{deuser2023sample4geo}.

\begin{figure}[!t]
\centering
\includegraphics[width=0.47\textwidth]{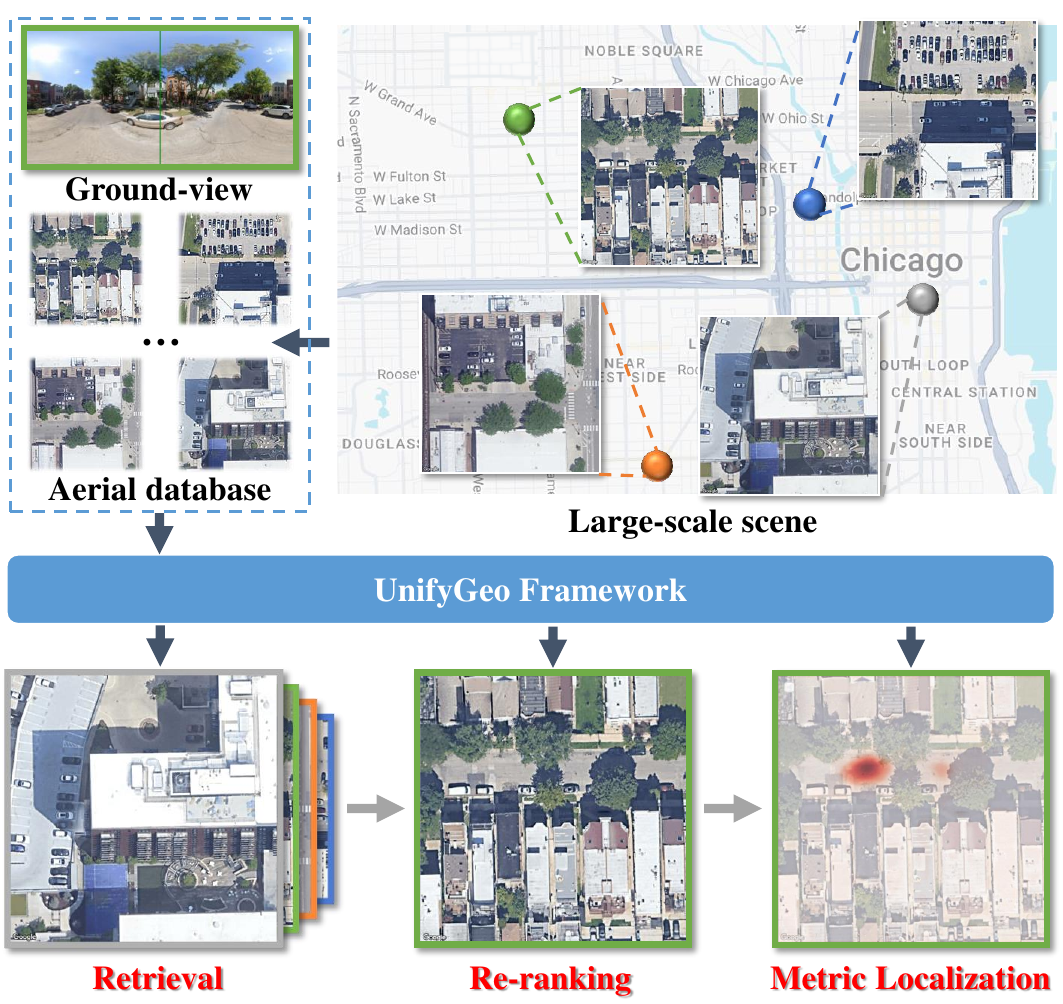}
\caption{
An illustration of our hierarchical geo-localization pipeline UnifyGeo for large-scale scenarios. 
Given a ground-level query image and an aerial image database covering the target area, our pipeline consists of three steps: 
(1) retrieval, which searches the entire database for potential aerial image candidates; 
(2) re-ranking, which selects the aerial image most likely to cover the query location
and (3) metric localization, which precisely estimates the query location using the re-ranked aerial image as reference.
}
\label{fig_1:motivation}
\end{figure}

Existing methods \cite{cai2019ground, shi2020optimal,  ye2025cross, hu2018cvm, zhu2022transgeo} commonly consider the CVGL task as a cross-view image retrieval problem, and train models to find better global image descriptors that bring matching ground-aerial pairs closer while pushing unmatching pairs far apart \cite{guo2022soft, song2026warpfree}.
However, the retrieval-based method only provides an approximation of the query’s geographic location to the center of the matched aerial image.
In many cases, the exact query location may not correspond to the center of any aerial image in the database \cite{zhu2021vigor}.
Recently, cross-view metric localization methods \cite{xia2022visual, shi2022beyond, lentsch2023slicematch} have been proposed to estimate the precise query location by registering the query view to a predefined satellite reference of the surrounding area.
However, acquiring a satellite reference image across large-scale scenarios is challenging, forcing these methods to rely on other coarse localization techniques
as a preliminary step.

Due to the complementary strengths of image retrieval and metric localization, a straightforward approach to achieve Large-scale Fine-grained CVGL (LF-CVGL) is to use image retrieval for obtaining an aerial image covering the query location, followed by a metric localization step for precise localization.
However, simply cascading standalone retrieval and metric localization models may suffer from insufficient task coordination and additional training overhead.
Although certain methods, such as \cite{zhu2021vigor, li2023patch}, have attempted to construct a unified network for joint task training, the coordination mechanisms of these two tasks still remain under-explored.
Specifically, the retrieval and metric localization tasks have different requirements for feature representation granularity, while existing methods \cite{zhu2021vigor, li2023patch} attempt to train a single representation that simultaneously satisfies both requirements, leading to conflicts in feature learning and reduced positioning performance.

In this paper, we propose UnifyGeo, a novel unified framework that integrates both retrieval and metric localization techniques to perform LF-CVGL. 
Unlike previous methods \cite{zhu2021vigor, li2023patch} that rely solely on global descriptors, UnifyGeo incorporates both global descriptors and detailed features for enhanced performance. 
Specifically, we design a unified learning strategy composed of \sz{a shared feature encoder with task-specific heads and a multi-task loss function.}
The shared encoder extracts uniform features for both retrieval and metric localization, effectively constructing task associations. 
On that basis, the multi-task loss function optimizes the encoder by simultaneously supervising global and detailed feature representations, promoting multi-granularity encoding that better supports both retrieval and metric localization. We also propose a feature aggregator to alleviate conflicts in representation learning \textcolor{black}{and improve the compatibility between retrieval and metric localization during joint optimization.}
Furthermore, we introduce a re-ranking mechanism guided by a dedicated loss function to refine the selection of aerial images from retrieval candidates. 
This ensures reliable reference data for metric localization, and further bridges the gap between retrieval and metric localization tasks.
The complete hierarchical geo-localization pipeline is illustrated in Fig.~\ref{fig_1:motivation}.

Overall, our contributions are as follows:

\begin{itemize}

\item We propose UnifyGeo, a novel hierarchical framework that integrates retrieval and metric localization into a unified network for LF-CVGL, operating in a coarse-to-fine manner.

\item We propose a unified learning strategy that integrates multi-granularity representations to support our geo-localization pipeline, together with a feature aggregator that alleviates representation conflicts \textcolor{black}{and improves the compatibility between retrieval and metric localization during joint optimization.}
\item We introduce a re-ranking strategy guided by a dedicated loss function to refine image retrieval results and provide better references for metric localization, further enhancing the positioning accuracy.

\item Extensive experiments demonstrate that UnifyGeo significantly outperforms existing methods on LF-CVGL, and our retrieval and metric localization units within UnifyGeo achieve comparable or better performance than state-of-the-art individual techniques.

\end{itemize}

\section{Related Work} % 从两个独立任务方向上总结问题，突出层级式定位。

\subsection{Cross-view Image Retrieval}
Lin et al. \cite{lin2015learning} and Workman et al. \cite{workman2015wide} introduce the first approaches that use image embeddings to address cross-view geo-localization task as a retrieval problem. 
Compared to earlier approaches using handcrafted features \cite{bansal2011geo, viswanathan2014vision}, they demonstrate the superiority and great potential of deep learning in this domain.  
Both approaches employ standard fully connected layers to aggregate local CNN features for retrieval tasks. 
To improve aggregation strategies, Hu et al. \cite{hu2018cvm} introduce learnable NetVLAD layers \cite{arandjelovic2016netvlad} to map CNN features to global descriptors. 
However, direct aggregation of CNN features fails to reduce the visual differences between cross-view images. 
To address this, Shi et al. \cite{shi2019spatial} apply a polar coordinate transformation to convert aerial images into a format visually similar to ground views, enabling the network to learn more consistent descriptors.
\textcolor{black}{Li et al. \cite{li2023multi} further enhance cross-view features with a multi-scale attention encoder that captures contextual information across scales.} 
Lin et al. \cite{lin2022joint} propose learning view-invariant keypoints to eliminate the domain gap at the feature level. 
Although these transformations improve retrieval performance by narrowing the domain gap, their effectiveness is primarily limited to one-to-one, center-aligned cross-view image matching.

To evaluate cross-view geo-localization in a more realistic setting, Zhu et al. \cite{zhu2021vigor} introduce the concept of semi-positive aerial images, which cover the same ground scene from different, non-central viewpoints. 
In response, some approaches have shifted away from coordinate transformation techniques and leverage novel architectures.
Zhu et al. \cite{zhu2022transgeo} and Yang et al. \cite{yang2021cross} are among the first to utilize the global information modeling capabilities of transformers to learn discriminative representations. 
Furthermore, Zhang et al. \cite{zhang2023cross} propose a novel approach that decouples content features from spatial layouts, using transformer layers to focus on spatial relationships. 
Additionally, Zhu et al. \cite{zhu2023simple} introduce a new architecture based on Multi-Head Attention (MA), specifically designed for geo-localization tasks.
Deuser et al. \cite{deuser2023sample4geo} simplify the dual-branch architecture into a siamese feature encoder, enhancing the representation power of the global descriptor with a ConvNeXt-base model and global hard sample mining strategies, achieving state-of-the-art retrieval performance.  
More recently, Shugaev et al. \cite{shugaev2024arcgeo} focus on optimizing for limited Field of View (FoV) cases.  
Li et al. \cite{li2024unleashing} explore unsupervised frameworks for real-world scenarios, although their retrieval performance lags behind supervised methods.

\subsection{Cross-view Metric Localization}

Given a ground-level query image and an aerial image covering its surroundings, metric localization aims to determine the precise position, and when the orientation is also estimated, this task is commonly referred to as pose estimation.
In \cite{zhu2021vigor}, Zhu et al. introduce the first dataset to evaluate this task and propose to regress query position by reusing global descriptors. 
Subsequently, Xia et al. \cite{xia2022visual} formulate the task as a multi-class classification problem and further improve the localization performance using a hierarchical matching approach \cite{xia2023convolutional}.
Chen et al. \cite{chen2024metric} also utilize matching-based techniques, generating location heatmaps for precise localization in lunar exploration.
Instead of letting the network automatically learn matching patterns, Lentsch et al. \cite{lentsch2023slicematch} pre-generate multiple hypothesized pose templates and create a set of map representations for the aerial images based on these templates. 
The final ground pose is determined by matching query features to those from the map representation set. 
Shi et al. \cite{shi2022beyond} propose an iterative framework for pose estimation, which projects aerial image features into the ground-view perspective via homography and refines the query pose using the Levenberg-Marquardt algorithm \textcolor{black}{\cite{levenberg1944method, marquardt1963algorithm}}. 
More recently, researchers have utilized Bird's Eye View (BEV) transformations, achieved through Vision Transformers \cite{fervers2023uncertainty} or spherical transforms \cite{wang2024fine}, to close the domain gap between ground and aerial views.
\textcolor{black}{The resulting BEV and aerial feature maps are densely matched for pose estimation.}
In contrast to existing metric localization methods, which are limited to localizing queries within a local area, our hierarchical geo-localization framework enables accurate geo-localization over large-scale scenarios. 

\subsection{Large-scale Fine-grained Cross-view Geo-localization}
To achieve \textcolor{black}{fine-grained cross-view geo-localization at large scale}, Zhu et al. \cite{zhu2021vigor} propose a model that generates global descriptors for cross-view pairs, using them for both image retrieval and location regression.
However, these global descriptors lack fine-grained scene information, limiting localization accuracy \cite{xia2022visual}.
To complement global descriptors with finer-grained cues, PaSS-KD \cite{li2023patch} uses multi-scale patch-level similarity as auxiliary supervision. 
In contrast, we decouple the representation requirements of retrieval and metric localization by learning multi-granularity features with task-specific heads, instead of reusing global descriptors. 
Different from the training-only local branch in PaSS-KD, our detailed features encode fine-grained spatial information and are explicitly used during inference for localization and candidate re-ranking, enabling higher localization accuracy and more reliable aerial reference selection. 

\section{METHODOLOGY}
In this section, we first explain the task of LF-CVGL (Section \uppercase\expandafter{\romannumeral3}.A) and our hierarchical geo-localization pipeline (Section \uppercase\expandafter{\romannumeral3}.B). We then introduce the key components of our proposed framework UnifyGeo (Section \uppercase\expandafter{\romannumeral3}.C) and the re-ranking \sz{mechanism} (Section \uppercase\expandafter{\romannumeral3}.D). Finally, we describe the training process of UnifyGeo in detail (Section \uppercase\expandafter{\romannumeral3}.E).

\subsection{Problem Formulation} 

Considering an aerial database $\mathcal{A} = \{I^1_a, \ldots, I^N_a\}$ of $N$ aerial/satellite images, each covering a square region associated with a different geo-location. The $N$ square regions form a complete map representation that covers all possible query locations. With database $\mathcal{A}$ as reference, given a ground-level query image $I_g$, we aim to estimate the geographic location, $\hat{\textbf{P}} = (\hat{x}, \hat{y})$, of camera that took $I_g$. Here, $(\hat{x}, \hat{y})$ is the image coordinate in a geo-tagged aerial image of known spatial resolution that covers the query image's camera location.

\subsection{Proposed Hierarchical Pipeline for LF-CVGL} 

\begin{figure*}[!t]
\centering
\includegraphics[width=0.95\textwidth]{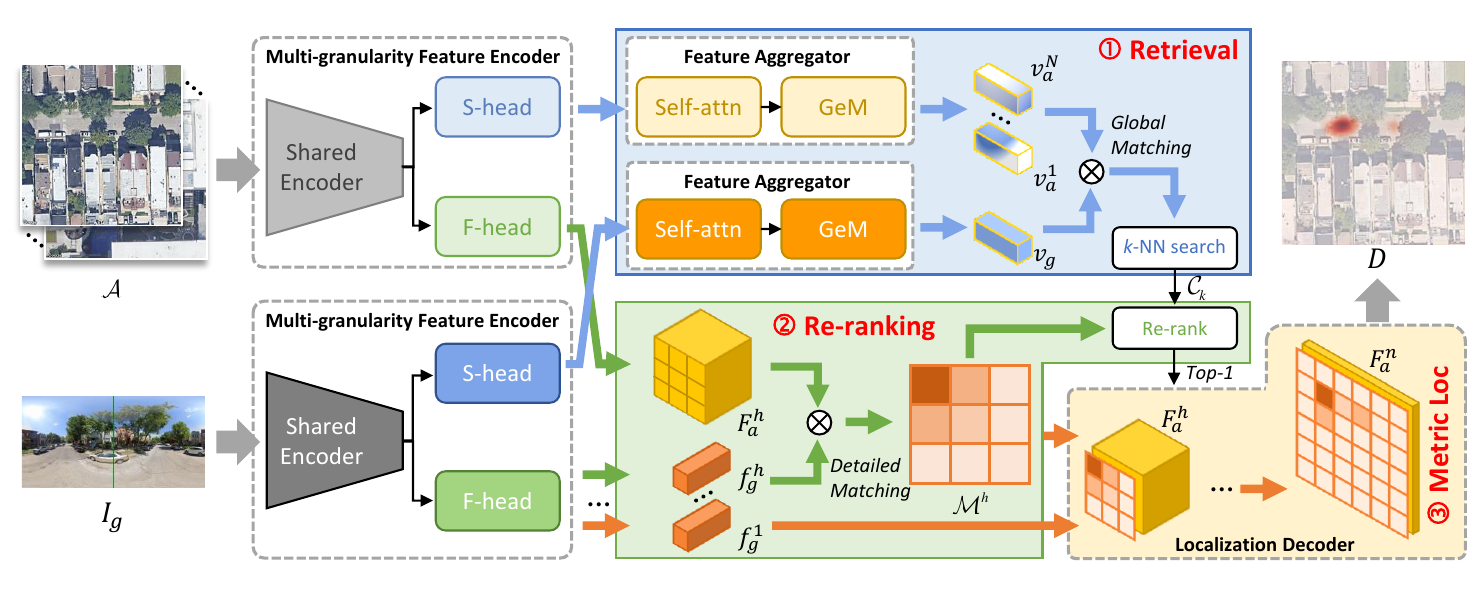}
\caption{
Overview of our proposed UnifyGeo framework. 
It consists of separate multi-granularity feature encoders and feature aggregators for the ground and aerial branches, as well as a localization decoder. Given a ground-level query image $I_g$ and an aerial database $\mathcal{A}$, we first perform cross-view image retrieval (Retrieval) to obtain $\mathcal{C}_k$, a set of $k$ aerial image candidates. These candidates are then re-ranked to select the top-1 matched aerial reference. Finally, cross-view metric localization (Metric Loc) is performed to produce a discrete probability distribution $D$, which is overlaid on the aerial reference to indicate the likely location of $I_g$, with deeper red denoting higher probability.
}
\label{fig_2:pipeline}
\end{figure*}

\sz{
Existing LF-CVGL methods typically follow a sequential pipeline of retrieval followed by metric localization, where query coordinates are regressed by reusing global descriptors.
However, such image-level representations may lack the fine-grained spatial information required for precise metric localization. 
Unlike PaSS-KD~\cite{li2023patch}, which uses patch-level and multi-scale cues as auxiliary supervision to enhance global descriptors, UnifyGeo learns multi-granularity representations and explicitly exploits detailed features during inference. 
These detailed features support both metric localization and candidate re-ranking, thereby refining retrieval results and providing more reliable aerial references for subsequent localization. 
As illustrated in Fig.~\ref{fig_2:pipeline}, we construct a coarse-to-fine hierarchical framework, which is summarized as follows.}

\textbf{Prior Retrieval.} 
To quickly narrow the geo-localization scope, image retrieval is performed by matching the global descriptors between the query image and the aerial database. 
Formally, given \sz{the $L2$-normalized} global descriptors $v_g$ and $v_a^i$ of $I_g$ and $I_a^i \in \mathcal{A}$, the retrieval task can be formulated as:
\textcolor{black}{
\begin{equation}
\mathcal{C}_k =
\operatorname*{arg\,top\text{-}k}_{I_a^i \in \mathcal{A}}
\operatorname{sim}\left(v_g,v_a^i\right),
\label{retrieval}
\end{equation}
where $\operatorname{sim}(\cdot,\cdot)$ denotes cosine similarity, and
$\operatorname{arg\,top\text{-}k}$ returns the $k$ aerial images with
the highest similarity scores, ordered in descending order.
Since the descriptors are L2-normalized, their cosine similarity can
be computed by the dot product. This retrieval can be efficiently
implemented using $k$-nearest-neighbor search. The resulting ranked
list of aerial candidates is denoted as $\mathcal{C}_k$.}

\textbf{Re-ranking and Location Estimation.} 
To accurately predict the query location, we need to identify the
geographically matched candidate in $\mathcal{C}_k$ that covers it.
To this end, we first compute a detailed matching score for each candidate
in $\mathcal{C}_k$ and combine it with the corresponding global retrieval
score. Thus, re-ranking considers both global semantic consistency and
fine-grained spatial correspondence. The aerial image with the highest
combined score is then selected as the reference for metric localization.
Based on the selected aerial image, the localization decoder generates an
$L\times L$ discrete probability distribution of the query location over
the aerial image. Finally, the maximum a posteriori (MAP) pixel in this
distribution is taken as the final predicted query location
$\hat{\mathbf{P}}$.

\subsection{UnifyGeo Architecture} 

UnifyGeo consists of three primary components: the multi-granularity feature encoder, the feature aggregator, and the localization decoder, as illustrated in Fig.~\ref{fig_2:pipeline}. 
The multi-granularity feature encoder extracts two distinct feature maps, each being tailored to address specific tasks.
The feature aggregator processes one feature map to generate a global descriptor for image retrieval, while the other feature map
is used to extract detailed features for both re-ranking\footnote{
\sz{
For clarity, we first introduce the detailed feature matching process in the localization decoder, and then explain how the same initial matching matrix is reused for re-ranking. 
This ordering is for exposition only and does not change the inference pipeline.
}
} and metric localization.

\textbf{Multi-granularity Feature Encoder.} 
Due to the natural differences in image characteristics across views, our feature encoder is a pseudo-Siamese network with two branches for ground and aerial views, \textcolor{black}{where the two branches have the same architecture but do not share weights}. 
Within each branch, we introduce two distinct processing pathways that share a single encoder for uniform feature computation, \textcolor{black}{followed by task-specific heads to adapt the features to global retrieval and fine-grained localization, respectively}.

Formally, given a ground query image $I_g$ and an aerial image $I_a \in \mathcal{A}$ with sizes $H \times W \times 3$ and $L \times L \times 3$, the shared encoders of the two branches first map them into feature maps $\{F_g^i\}$ and $\{F_a^i\}$, where $i \in \{\textcolor{black}{1, \ldots, n}\}$ and $n$ corresponds to the number of \textcolor{black}{shared} encoder stages.
Then, for the aerial view branch, the multi-granularity feature extractor heads act on top of the last stage's outputs $F_a^{\textcolor{black}{n}}$.
Specifically, the fine-grained extractor head (denoted as F-head in Fig.~\ref{fig_2:pipeline}) further enriches the image representation and produces a detailed feature map $F_a^{\textcolor{black}{h}}$ with dimension $L^{\prime} \times L^{\prime} \times C$.
The semantic extractor head (denoted as S-head in Fig.~\ref{fig_2:pipeline}) also takes $F_a^{\textcolor{black}{n}}$ as its input to extract semantic-level latent features $G_a^{\textcolor{black}{h}}$, which are then processed by our feature aggregator to generate a global descriptor $v_a$ for the aerial image.
Similarly, the ground image branch obtains a detailed feature map $F_g^{\textcolor{black}{h}}$ of size $H^{\prime} \times W^{\prime} \times C$ and a global descriptor $v_g$ in the same way.
Finally, the global descriptors $v_g, v_a$ are used for retrieval, while the multi-scale feature maps $\{ F_a^i \mid i \in \{\textcolor{black}{1, \ldots, n}\} \}$ \textcolor{black}{together with $F_a^h$} and $F_g^{\textcolor{black}{h}}$ are used as input for both re-ranking and metric localization.

\textbf{Feature Aggregator.} 
In UnifyGeo, we design a feature aggregator to extract global descriptors while \textcolor{black}{alleviating the representation conflict} in the aerial branch. 
This conflict arises from the divergent learning objectives of retrieval and metric localization tasks, both of which operate on the same feature maps produced by the shared encoder.
Specifically, metric localization requires aligning local regions in the feature map with detailed ground-view descriptors, whereas retrieval demands abstracting the entire feature map into a vector similar to the ground-view global descriptor. 
To \textcolor{black}{alleviate} this conflict, we introduce a self-attention mechanism into the aggregator.
This mechanism dynamically assigns weights to different features based on their correlation, selectively highlighting the most critical patterns for retrieval while avoiding ineffective aggregation from conflicting features. 
We further aggregate these critical patterns using generalized mean (GeM) pooling \cite{radenovic2018fine, ng2020solar}, which adjusts the pooling process with learnable parameters to preserve discriminative information in the feature map.
Moreover, as not all information in ground images is present in the aerial images (e.g., sky) \cite{lentsch2023slicematch}, we also employ the feature aggregator in the ground branch to facilitate the collection of critical representations that can enable contrastive learning with aerial global descriptors.

Concretely, given the features $\mathbf{G}$ produced by the semantic extractor head, $\mathbf{G}$ can either be $G_a \in \mathbb{R}^{L' \times L' \times C}$ or $G_g \in \mathbb{R}^{H'\times W'\times C}$. 
Each vector $G_{i, j}$ with dimension $C$ located at $\left(i, j\right)$ in the spatial direction can be considered as a region descriptor corresponding to a local area of the source image. 
To capture key semantic information from the global context, the self-attention module first maps region descriptors into a latent space, generating query ($\mathbf{Q}$), key ($\mathbf{K}$), and value ($\mathbf{V}$) vectors.
Then, the process can be represented as follows: 
\begin{equation}
\begin{aligned}
\mathbf{Q} &= LN(\mathbf{G})\mathbf{W}^{q},\quad
\mathbf{K} = LN(\mathbf{G})\mathbf{W}^{k},\quad
\mathbf{V} = \mathbf{G}\mathbf{W}^{v},\\
\mathbf{\Phi}
&=
\mathbf{G}
+
\operatorname{softmax}
\left(
\frac{\mathbf{Q}\mathbf{K}^{T}}{\sqrt{\lambda}}
\right)
\mathbf{V}\textcolor{black}{\mathbf{W}^{o}},
\end{aligned}
\label{feature_aggregator}
\end{equation}
\textcolor{black}{where the spatial dimensions of $\mathbf{G}$ are flattened before
attention and restored in $\mathbf{\Phi}$ afterward.
$\mathbf{W}^{q}$, $\mathbf{W}^{k}$, and $\mathbf{W}^{v}$ reduce the
channel dimension from $C$ to $\lambda=C/2$, while $\mathbf{W}^{o}$
restores it to $C$. $LN(\cdot)$ denotes LayerNorm, and
$1/\sqrt{\lambda}$ is the attention scaling factor.}
Note that we do not employ multi-head attention, 
this decision allows us to compute attention based on the full semantics of the region descriptors.
Furthermore, we omit the feed-forward network (FFN) commonly found in standard transformer blocks, as it offers only marginal performance improvement at a significant parameter cost \cite{zhu2023simple}. Finally, the enhanced feature map $\mathbf{\Phi}$
\textcolor{black}{(clamped element-wise to a minimum value of
$\epsilon=10^{-6}$)} is fed into GeM to generate a single global descriptor $v$.
\begin{equation}
v=\operatorname{GeM}(\mathbf{\Phi},p)
=
\left(
\frac{1}{H_{\Phi}W_{\Phi}}
\sum_{i=1}^{H_{\Phi}}
\sum_{j=1}^{W_{\Phi}}
\mathbf{\Phi}_{i,j}^{p}
\right)^{\frac{1}{p}},
\end{equation}
where $p$ is a learnable parameter that represents the generalized mean power, $H_{\Phi}$ and $W_{\Phi}$ denote the spatial resolution of $\mathbf{\Phi}$. 
Given the $L2$-normalized global descriptors from the ground and aerial branches, the retrieval step is then performed to obtain $\mathcal{C}_k$, a set of $k$ aerial image \sz{candidates}.

\begin{figure}[!t]
\centering
\includegraphics[width=0.49\textwidth]{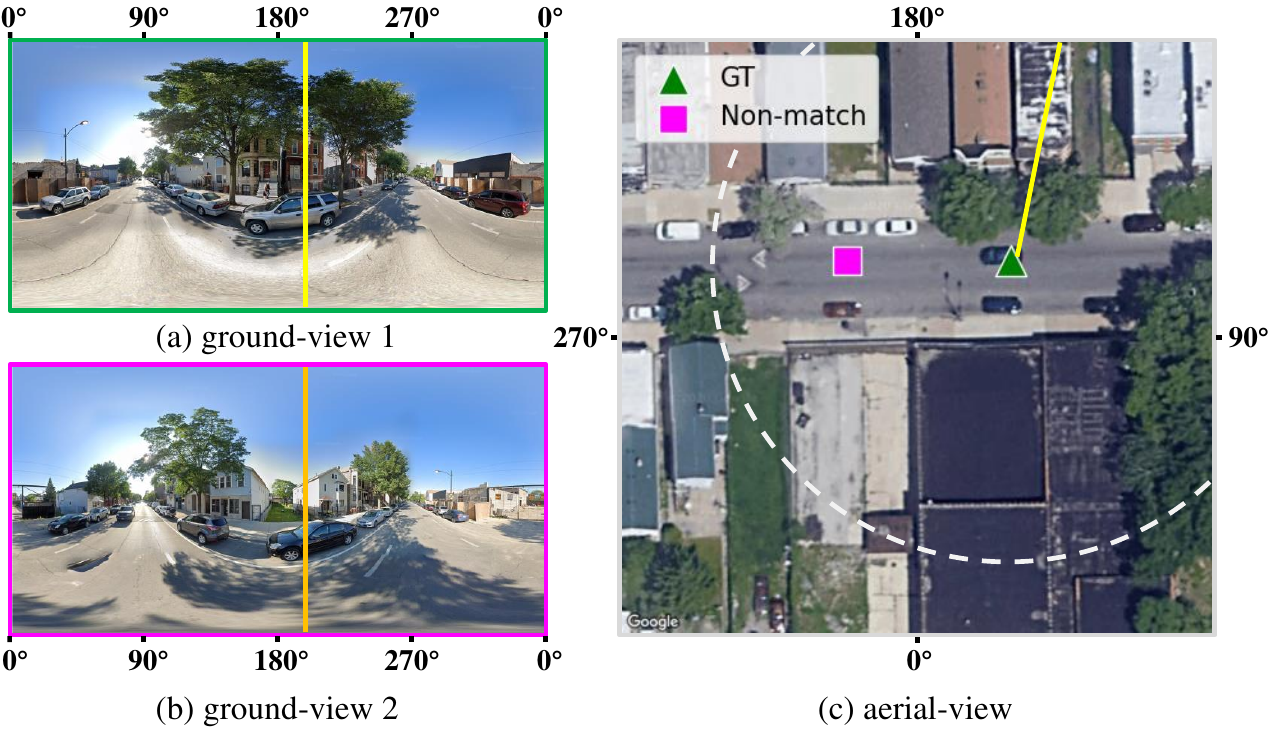}
\caption{
Illustration of the geometric correspondence between aerial and ground views.
The \textcolor{Green}{\textbf{green triangle}} on the aerial
image (c) corresponds to the ground-truth location of image (a), marked as GT, with a roughly matching field of view indicated by the white dashed line in (c).
In contrast, the \textcolor{Magenta}{\textbf{magenta square}} on (c) does not match (a), with its corresponding ground-view image shown in (b).
Notably, the same azimuthal direction (yellow bar) in (a) and (c) displays the same scene, featuring trees and a red facade building.
Conversely, the same azimuthal direction (orange bar) in (b) depicts a different scene, showing a white building.
}
\label{fig_3:spatial cor}
\end{figure}

\textbf{Localization Decoder.} 
Given the detailed features of both the aerial candidate from $\mathcal{C}_k$ and $I_g$, the localization decoder predicts the possible location of $I_g$ for metric localization by generating a dense probability distribution over the aerial candidate. 
To obtain a high-resolution localization distribution, we employ a hierarchical matching strategy inspired by \cite{xia2023convolutional}.  
This strategy involves iteratively refining the matching process between the ground descriptor and local regions in the aerial feature map, starting from a coarse resolution and progressively increasing the spatial detail.
At each iteration, the matching score is used to inform the upsampling of the aerial features to a higher resolution.

Specifically, for the aerial branch, multi-scale features $\{ F_a^i \mid i \in \{\textcolor{black}{1, \ldots, n}\} \}$ \textcolor{black}{and $F_a^h$} from the encoder and fine-grained extractor head are used for hierarchical matching.
Corresponding detailed ground descriptors $\{ f_g^i \mid i \in \{\textcolor{black}{1, \ldots, n}\} \}$ \textcolor{black}{and $f_g^h$} are generated through $n+1$ projections of the ground detailed feature map $F_g^{\textcolor{black}{h}} \in \mathbb{R}^{H' \times W' \times C}$.
To enhance the generation of detailed ground descriptors, we follow \cite{lentsch2023slicematch, xia2023convolutional} to incorporate geometric constraints into the projector design, leveraging the fact \cite{shi2019spatial} that pixels in the same azimuthal direction in aerial images typically correspond to vertical columns in ground-view images, as shown in Fig.~\ref{fig_3:spatial cor}. 
By aggregating column-wise information from the ground feature maps, our projectors can capture scene details more effectively.
Each projector comprises a $1 \times 1$ convolutional layer that reduces the feature channels, followed by a column-wise fully connected \sz{(FC)} operation that summarizes the vertical information, thereby collapsing the column dimension from $H^\prime$ to 1. The resulting feature map is then reshaped into a compact 1D vector, yielding a detailed ground descriptor.
\sz{Algorithm~\ref{alg:column_collapse} provides a pseudocode implementation.}

\begin{algorithm}[t]
\caption{\textcolor{black}{Detailed Ground Descriptor Projection}}
\label{alg:column_collapse}
\textcolor{black}{\textbf{Input:} Ground detailed feature map $F_g^h \in \mathbb{R}^{H^\prime \times W^\prime \times C}$.}\\
\textcolor{black}{\textbf{Output:} Detailed ground descriptors $\{f_g^l \mid l \in \{1,\ldots,n,h\}\}$.}
\begin{algorithmic}[1]
\State \textcolor{black}{Let $\mathcal{S}=\{1,\ldots,n,h\}$ denote the target hierarchy levels.}
\For{\textcolor{black}{each level $l \in \mathcal{S}$}}
    \State \textcolor{black}{Set $C_l^\prime=C_l/W^\prime$, where $C_l$ is the channel dimension of the corresponding aerial feature map $F_a^l$.}
    \State \textcolor{black}{$F_c^l = \operatorname{Conv}_{1\times1}^l(F_g^h) \in \mathbb{R}^{H^\prime \times W^\prime \times C_l^\prime}$.}
    \For{\textcolor{black}{$w = 1$ to $W^\prime$}}
        \State \textcolor{black}{$p_w^l = \operatorname{FC}^l(F_c^l[:,w,:]) \in \mathbb{R}^{1 \times C_l^\prime}$.}
    \EndFor
    \State \textcolor{black}{$P_g^l = \operatorname{Concat}(p_1^l,\ldots,p_{W^\prime}^l) \in \mathbb{R}^{1 \times W^\prime \times C_l^\prime}$.}
    \State \textcolor{black}{$f_g^l = \operatorname{Reshape}(P_g^l) \in \mathbb{R}^{W^\prime C_l^\prime}$.}
\EndFor
\State \textcolor{black}{Return $\{f_g^l \in \mathbb{R}^{C_l} \mid l \in \mathcal{S}\}$.}
\end{algorithmic}
\end{algorithm}

Next, hierarchical detailed feature matching starts from the lowest resolution at level $\textcolor{black}{h}$ and moves toward \textcolor{black}{level $1$}. 
The initial score matrix $\mathcal{M}^{\textcolor{black}{h}}$ is computed by matching $f_g^{\textcolor{black}{h}}$ to $F_a^{\textcolor{black}{h}}$, with LayerNorm and $L2$-normalization being applied.
Next, $\mathcal{M}^{\textcolor{black}{h}}$ is concatenated with the $L2$-normalized $F_a^{\textcolor{black}{h}}$ and used to guide the upsampling of aerial features through deconvolution. 
Skip connections are added from previous aerial features for better scene layout information, followed by convolution to form the next-level matching feature maps. 
This process continues until \textcolor{black}{level $1$} is reached, generating matching scores \textcolor{black}{$\{ \mathcal{M}^{h}, \mathcal{M}^{n}, \ldots, \mathcal{M}^{1} \}$}. 
The highest-resolution score matrix with corresponding \textcolor{black}{level-1} feature maps \textcolor{black}{is} upsampled to the target resolution $L \times L \times 1$ , and a softmax activation further converts it into a discrete distribution $D$. 
The maximum a posteriori (MAP) pixel location $(\hat{x}, \hat{y})$ in $D$ yields the final location prediction.

\vspace{-0.3cm}
\subsection{Re-ranking Mechanism} 

In addition to retrieval and metric localization, we introduce a re-ranking mechanism that is essential to our hierarchical geo-localization pipeline.
The purpose of re-ranking is to select the correct aerial image that covers the query location from the top-$k$ retrieval candidates, thereby preventing erroneous location estimation from non-matching aerial images.
Thanks to joint training with detailed features, UnifyGeo enables fine-grained comparisons between the query view and each retrieved aerial candidate for re-ranking.

Specifically, for each of the top-$k$ aerial candidates $I_a^{t} \in \mathcal{C}_k$, we perform detailed feature matching with the query view to compute an initial score matrix $\mathcal{M}_t^{\textcolor{black}{h}}$, using the same matching strategy as in the metric localization step.
\sz{For efficiency and consistency, $\mathcal{M}_t^{h}$ is computed only once for each candidate and then reused by the re-ranking process and by the localization decoder to support metric localization.}
The re-ranking process is then formulated as:
\begin{equation}
\label{eq:re-ranking-sum}
I_a^* = \mathop{\arg\max}_{I_a^{t} \in \mathcal{C}_k} \left( \sz{\operatorname{sim}(v_g, v_a^t)} + \max_{(i, j)} \mathcal{M}_t^{\textcolor{black}{h}}(i, j) \right),
\end{equation}
where \sz{$\operatorname{sim}(v_g, v_a^t)$} denotes the prior retrieval score of candidate $I_a^{t}$.
Note that each value in the score matrix $\mathcal{M}_t^{\textcolor{black}{h}}$ reflects the similarity between a specific region in the aerial image and the query ground-view, derived through detailed feature matching.
For the aerial candidate that truly covers the query location, the highest-scoring position in its score matrix should correspond to the query location, as the detailed features here describe the same geographic area as the detailed query descriptor. 
We perform re-ranking by combining each candidate’s maximum score from the initial score matrix with its retrieval score, thereby balancing overall layout and scene details. 
\sz{A sensitivity analysis of different fusion weights for these two scores is provided in Section~IV-G.}
\sz{Finally,} the aerial image with the highest combined score is selected as our re-ranking result.

\subsection{Training Strategy for UnifyGeo}

Our method aims to jointly retrieve the geographically matched aerial image for a given ground-level query image and accurately estimate the query location
using the matched aerial image as the reference.
To further improve retrieval performance, we incorporate a re-ranking process into the training pipeline.
The multi-task loss function is defined as:
\begin{equation}
\label{overall_loss1}
\mathcal{L} = \mathcal{L}_D + \alpha\mathcal{L}_G + \beta\mathcal{L}_M + \gamma\mathcal{L}_R,
\end{equation}
where $\alpha$, $\beta$ and $\gamma$ are hyperparameters during training.
\sz{Here, $\mathcal{L}_D$ supervises the localization distribution, while $\mathcal{L}_G$, $\mathcal{L}_M$, and $\mathcal{L}_R$ denote the retrieval, hierarchical matching, and re-ranking losses, respectively. 
All loss terms are detailed below.}

\textbf{Retrieval Loss.}  For image retrieval, we train the global descriptors extracted from both the aerial and ground branches using contrastive learning. 
To fully leverage the negative examples within the batch, we follow \cite{deuser2023sample4geo} to utilize the InfoNCE loss in a symmetric manner.
This loss is designed to push the matched pairs closer in latent space and pull the unmatched pairs farther apart. The InfoNCE loss used for retrieval is formulated as:
\begin{equation}
\label{infonce}
\mathcal{L}_{\text {InfoNCE }}=-\log \frac{\exp \left(q \cdot r_{+} / \tau\right)}{\sum_{i=1}^N \exp \left(q \cdot r_i / \tau\right)},
\end{equation}
where $q$ represents the encoded query, $\{r_i\}$ denotes the encoded
reference samples containing the sole positive $r_{+}$ and the remaining
negatives, and $\tau$ is a learnable temperature parameter.
For the symmetric InfoNCE loss, $q$ can either be a ground image or an aerial image, depending on the direction of the information flow. 
Regardless of the direction, there is only one positive sample, $r_{+}$, that corresponds to $q$. 
The loss $\mathcal{L}_G$ is the average of the InfoNCE losses for both directions. 

\textbf{Localization Loss.} For metric localization, we apply $\mathcal{L}_D$ for location estimation, and $\mathcal{L}_M$ for the detailed matching process.
Specifically, we follow \cite{xia2023convolutional} to formulate the localization problem as a multi-class classification and apply a cross-entropy loss on the discrete probability distribution $D$.
The ground truth location is represented by a discrete distribution $D_{gt}$ with 2D Gaussian label smoothing of same size $L \times L$ with $D$.
Then, $\mathcal{L}_D$ is computed by:
\begin{equation}
\label{localization loss}
\mathcal{L}_D=-\sum_{i=1}^L \sum_{j=1}^L D_{gt}^{i, j} \log D^{i, j}.
\end{equation}

Since the localization probability distribution $D$ is generated step by step by the matching score, we apply \textcolor{black}{$\mathcal{L}_M = \sum_{l \in \{1, \ldots, n, h\}} \mathcal{L}_{M^{l}}$} on each matching score matrix of different levels.
At level $l$, $\mathcal{L}_{M^l}$ is used to encourage the aerial region descriptors $F_{a}^l(i, j)$ for locations close to the ground truth positions to match the detailed ground descriptor $f_g^l$.
As for the ground truth of level $l$, we max-pool $D_{gt}$ from $L \times L$ spatial dimensions to $L_l \times L_l$ resolution and renormalize it to generate weights $w^{l}_{i,j}$.
The loss $\mathcal{L}_{M^l}$ is finally defined as a weighted sum $\mathcal{L}_{M^l} = \sum_{i,j} w^{l}_{i,j} \mathcal{L}'_{M^l}(i,j) $ of InfoNCE loss based on cosine similarity,
\begin{equation}
\label{match loss}
\mathcal{L}^{\prime}{ }_{M^l}(i, j)=-\log \frac{\exp \left(\operatorname{sim}\left(f_g^l, F_{a}^l(i, j)\right) / \tau\right)}{\sum_{i^{\prime}, j^{\prime}} \exp \left(\operatorname{sim}\left(f_g^l, F_{a}^l(i^{\prime}, j^{\prime}) \right) / \tau\right)}.
\end{equation}

\textbf{Re-ranking Loss.} 
In our hierarchical geo-localization pipeline, the detailed feature matching
matrix $\mathcal{M}^{h}$ plays a crucial role in the re-ranking process.
During metric localization, the loss $\mathcal{L}_{M^{h}}$ is applied to
$\mathcal{M}^{h}$ to ensure that this matrix effectively guides the generation
of the discrete probability distribution. This loss encourages the highest
matching responses to correspond to the region in $\mathcal{M}^{h}$ that
aligns with the ground camera's location. However,
$\mathcal{L}_{M^{h}}$ is constrained to a single aerial feature map with a
resolution of $L' \times L'$, whereas the re-ranking process requires detailed
matching score comparisons across multiple aerial images.

\textcolor{black}{To overcome this limitation, we introduce an additional supervision term
$\mathcal{L}_{R}$ on the detailed matching scores at level $h$. Specifically,
during training, we extend the scope of comparison by matching each detailed
ground descriptor $f^{h}_{g,b}$ against the feature maps
$\{F^{h}_{a,m}\}_{m=1}^{B}$ of all aerial images within a mini-batch of size
$B$. Let
$\mathcal{M}^{h}_{b,m}(i,j)
=\operatorname{sim}\!\left(f^{h}_{g,b},F^{h}_{a,m}(i,j)\right)$
denote the matching score between the $b$-th ground query and location $(i,j)$
in the $m$-th aerial feature map. Following Eq.~\ref{match loss},
$\mathcal{L}_{R}$ is defined as the weighted sum
$\mathcal{L}_{R}=\frac{1}{B}\sum_{b=1}^{B}\sum_{i,j}
w^{h}_{b,i,j}\mathcal{L}'_{R}(b,i,j)$, where
\begin{equation}
\mathcal{L}'_{R}(b,i,j)
=
-\log
\frac{
\exp\!\left(\mathcal{M}^{h}_{b,b}(i,j)/\tau\right)
}{
\displaystyle
\sum_{m=1}^{B}\sum_{i',j'}
\exp\!\left(\mathcal{M}^{h}_{b,m}(i',j')/\tau\right)
}.
\label{rerank_loss}
\end{equation}
Here, $b$ indexes the ground--aerial pairs, while $m$ indexes all
aerial images in the mini-batch.
This supervision encourages the ground-truth region in the paired aerial image to produce high matching responses relative to all aerial detailed feature maps, thereby improving
re-ranking effectiveness.}

\section{Experiments}
In this section, we first introduce the three large-scale cross-view geo-localization datasets, evaluation metrics, and our implementation details.
We then evaluate the localization accuracy of our UnifyGeo on the LF-CVGL task \textcolor{black}{and analyze its computational efficiency}. 
Next, we compare the performance of UnifyGeo with state-of-the-art methods on the retrieval and metric localization tasks separately, demonstrating the effectiveness of our pipeline's components.
Furthermore, we also provide qualitative results to show the differences between different methods \textcolor{black}{and discuss the limitations of our method}. 
Finally, we conduct an extensive ablation study \textcolor{black}{and architectural analysis} to validate our design choices and configurations.

\subsection{Datasets}

\textbf{The VIGOR dataset} \cite{zhu2021vigor} collects 90,618 aerial images covering 4 major cities in U.S.A, including New York City, San Francisco, Chicago, and Seattle.
105,214 street-view images with unique GPS locations are collected for query purposes. These data are balanced to ensure that each aerial image contains at most two positive street-view images. A positive pair means that the street-view image lies within the central quarter of the corresponding aerial image. To better simulate real-world conditions, for each positive pair, there are also three semi-positive aerial-view neighbors that cover regions of the street-view image, with the street-view image located around the periphery of these three aerial views. 
In addition to the ground truth of pairing relationships, the VIGOR dataset also provides the GPS coordinates of each query image and the offset relative to the center of the positive or semi-positive aerial image for meter-level localization evaluation. 
To better evaluate cross-view localization methods, 
the VIGOR dataset introduces two distinct evaluation protocols: same-area and cross-area splits. 
The same-area protocol assesses performance by using images from all available cities \textcolor{black}{while keeping the training and testing ground-aerial image pairs disjoint}, whereas the cross-area protocol evaluates cross-city generalization by training on images from New York and Seattle and testing exclusively on data from San Francisco and Chicago.
The resolution of aerial and street-view images in VIGOR is $640 \times 640$ and $1024 \times 2048$, respectively. 
To ensure a fair comparison, we follow \cite{deuser2023sample4geo, li2023patch} and \textcolor{black}{resize the street-view and aerial images to $384 \times 768$ and $384 \times 384$, respectively, while preserving their original aspect ratios}.

\textbf{CVUSA and CVACT} are two widely used datasets for evaluating the cross-view image retrieval task \cite{workman2015wide, liu2019lending}. 
Each dataset contains 35,532 cross-view pairs for training and 8,884 pairs for evaluation (test for CVUSA and validation for CVACT).
Additionally, CVACT provides 92,802 test pairs with accurate geo-tags, referred to as CVACT\_test, to assess the generalization ability of the evaluated model.
Unlike VIGOR, these two datasets warp the panoramas to align with the center of the aerial images and adopt one-to-one retrieval for evaluation. 
During the experiments, we standardize the resolution of the cross-view images from both datasets to a consistent format, specifically $140 \times 768$ for the street-view images and $384 \times 384$ for the aerial-view images. 

Notably, for the evaluation of LF-CVGL, we primarily use the VIGOR dataset, as it is, to our knowledge, the only cross-view dataset capable of assessing model performance across large-scale scenes while also supporting fine-grained localization evaluation.
In contrast, CVUSA and CVACT have coarse-grained paired labels, but their center-aligned cross-view image pairs enable precise geo-localization via successful top-1 retrieval. We therefore use these datasets as auxiliary evaluation benchmarks for LF-CVGL.
Other metric localization datasets, such as KITTI \cite{geiger2013vision} and Ford multi-AV \cite{agarwal2020ford}, support fine-grained evaluation but are restricted to local localization and are not suitable for large-scale scenarios.

\subsection{Evaluation Metrics}

In our experiments, we evaluate three related tasks: LF-CVGL, image retrieval, and metric localization.
For LF-CVGL, we \sz{follow the meter-level evaluation protocol of VIGOR~\cite{zhu2021vigor}}, which measures the percentage of query images localized within a \sz{specified distance threshold} from the ground-truth camera location. 
\sz{Specifically, the predicted offset within the retrieved aerial image is first converted into GPS coordinates according to the geo-reference of the aerial image. 
We then compute the distance in meters between the predicted GPS location and the ground-truth GPS location of the query camera, and a prediction is considered correct if this distance falls within the threshold.}
On the VIGOR dataset, we consider two \sz{localization thresholds}: high-precision $(1m)$, and coarse-precision $(10m)$, denoted as ``$R@1m$" and ``$R@10m$" respectively.
On the CVUSA and CVACT datasets, 
we report the LF-CVGL performance as the percentage of query images with the ground-truth aerial image ranked first in retrieval orders.
For image retrieval, we use the top-$k$ recall accuracy metric \cite{deuser2023sample4geo, zhu2022transgeo, shi2019spatial}, denoted as ``$R@k$".
A retrieval is considered successful if the ground-truth aerial image is ranked within the top-$k$ retrieved images for a given ground-level query image. 
On the VIGOR dataset, we also report the hit rate \cite{zhu2021vigor}, which considers top-1 accuracy with both positive and semi-positive samples as correct results.
For metric localization, we follow the evaluation protocol used in \cite{zhu2021vigor, xia2023convolutional} and report the mean and median errors (in meters) between predicted and ground-truth positions for all test image pairs.

\subsection{Implementation Details}

To balance encoding capability with parameter efficiency, the multi-granularity feature encoder in UnifyGeo is based on ConvNeXt-Tiny. 
We split ConvNeXt-Tiny (before the global average pooling layer) into two parts: the first three stages serve as the shared feature encoder, while Stage 4 is duplicated into two separate heads as the fine-grained feature extractor and the semantic feature extractor, without sharing parameters between these two extractors. \textcolor{black}{In the notation of Section~III-C, the shared encoder is indexed from Stage $1$ to Stage $n$ with $n=3$, while the duplicated Stage 4 outputs are denoted by the head superscript $h$.}
Then, we retain only the first down-sampling layer and the ConvNeXt block in Stage 4 for constructing the extractor heads.
In the feature aggregator, we halve the dimensions of $\mathbf{Q}$, $\mathbf{K}$, and $\mathbf{V}$, and then restore them before adding a residual connection to the input. 
This results in a global descriptor dimension of 768 after GeM pooling.
For the localization decoder, hierarchical detailed feature matching is performed at \textcolor{black}{4 levels}, with the lowest resolution of $12 \times 12$ at \textcolor{black}{the head level $h$}. 
The number of feature channels and the dimensionality of the corresponding detailed ground descriptors are equal at each level, specifically 96, 192, 384, and 768. 

For training, we follow \cite{deuser2023sample4geo} by applying a label smoothing of 0.1 in the retrieval InfoNCE loss. 
Consistent with \cite{deuser2023sample4geo} and \cite{zhang2023cross}, we also apply synchronized horizontal flipping and rotation of cross-view image pairs, along with data augmentation techniques such as color jitter, grid dropout, and coarse dropout, to enhance model generalization. 
Different loss weight combinations selected from $\{0.1, 1, 10, 100\}$ were tested for balancing $\mathcal{L}_D$, $\mathcal{L}_G$, $\mathcal{L}_M$, and $\mathcal{L}_R$, while maintaining comparable loss scales during training.
\textcolor{black}{We set $\alpha=100$, $\beta=10$, and $\gamma=1$ based on R@1m performance
on the official VIGOR same-area test split and keep these weights fixed for
all other evaluation settings and datasets.}
When computing $\mathcal{L}_R$, all other aerial images in the mini-batch are used as negatives for each query, and semi-positive aerial images in VIGOR are not explicitly excluded. Since semi-positive images only partially cover the query scene and usually place the query location near the boundary, they do not provide reliable high-confidence regions for fine-grained localization. They are therefore treated as negatives, similar to positions far from the ground-truth center in the positive aerial image that receive near-zero weights under the Gaussian localization target.
\sz{For optimization, all models are trained for 40 epochs with a batch size of 64. We use AdamW as the optimizer with a weight decay of 0.04. The initial learning rate is set to 4.5e-4 and follows a cosine decay schedule, with a linear warm-up over the first epoch. Gradient clipping is applied with a maximum value of 100.} 

\textcolor{black}{For inference, we retrieve and re-rank the top-$k$ aerial candidates.
Based on results on the official VIGOR same-area test split, we use $k=5$
by default and additionally report $k=3$ as an efficiency-oriented alternative
in Table~\ref{tab:efficiency_analysis}.
We set $\eta=0.5$ to equally weight the global retrieval and fine-grained
matching scores without dataset-specific tuning.
Sensitivity analyses of $k$ and $\eta$ are provided in Sections~IV.F and IV.G,
respectively.}

\begin{figure*}[!t] % 图1

    \centering
    \subfloat[VIGOR Same-area setting]{ % 这里的 caption 让 subfloat 自动生成 (a), (b)
        \includegraphics[width=0.49\textwidth]{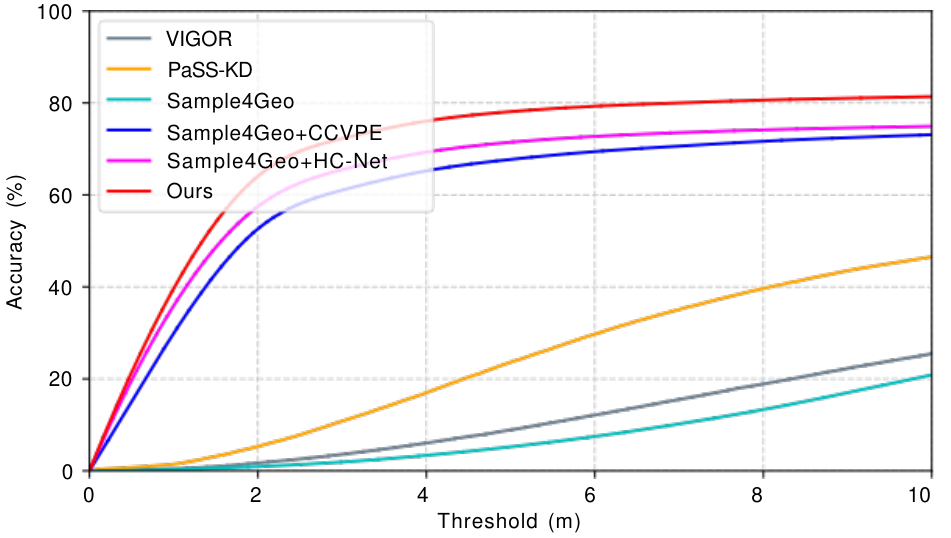}
        \label{fig:same_area_plot}}
    \subfloat[VIGOR Cross-area setting]{
        \includegraphics[width=0.49\textwidth]{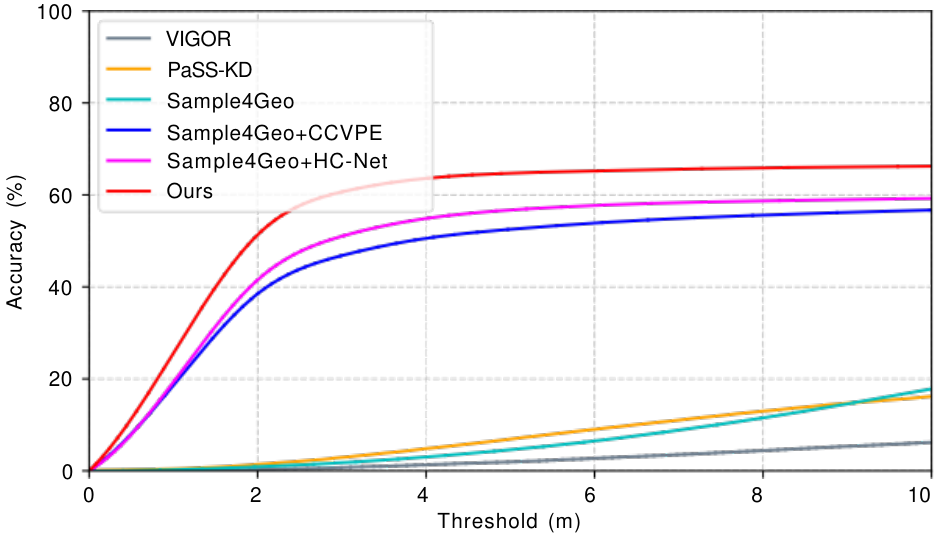}
        \label{fig:cross_area_plot}}
    \caption{Localization accuracy at various thresholds for the LF-CVGL task on the VIGOR dataset. Our method UnifyGeo is compared to the state-of-the-art methods and combined baselines on both the same-area (a) and cross-area (b) settings.}
    \label{fig_4: loc_experiment}

\end{figure*}

\begin{table}[t] % 表1
\centering
\caption{Comparison of meter-level localization accuracy with state-of-the-art methods for LF-CVGL on the VIGOR dataset. }
\label{tab1_loc_vigor}
\begin{tabular}{l|cc|cc} 
\hline\hline
\multirow{2}{*}{Methods} & \multicolumn{2}{c|}{Same-area}  & \multicolumn{2}{c}{Cross-area}   \\
                          & R@1m           & R@10m          & R@1m           & R@10m           \\ 
\hline
VIGOR \cite{zhu2021vigor} & 0.55           & 25.50          & 0.06           & 6.20            \\
PaSS-KD \cite{li2023patch}& 1.53           & 46.50          & 0.43           & 16.20           \\ 
Sample4Geo \cite{deuser2023sample4geo} & 0.24           & 20.84          & 0.23           & 17.81           \\
\hline
Sample4Geo+CCVPE          & 29.90          & 73.10          & 18.61          & 56.71           \\
Sample4Geo+HC-Net         & 35.92          & 74.91          & 19.28          & 59.23           \\
\textcolor{black}{UnifyGeo (ours)}             & \textbf{39.64} & \textbf{81.37} & \textbf{25.58} & \textbf{66.24}  \\
\hline\hline
\end{tabular}
\end{table}

\begin{table}[t] % 表2
\centering
\caption{Comparison of efficiency on the VIGOR same-area setting.}
\label{tab:efficiency_analysis}
\resizebox{\columnwidth}{!}{%
\begin{tabular}{lcccc}
\toprule
Methods & Params. $\downarrow$ & Time $\downarrow$ & Mem. $\downarrow$ & $R@1m$ $\uparrow$ \\
\midrule
Sample4Geo+HC-Net & 99.7M & \textbf{12.30ms} & 6.36GB & 35.92 \\
\textcolor{black}{UnifyGeo ($k=3$)} & \textcolor{black}{\textbf{57.7M}} & \textcolor{black}{29.20ms} & \textcolor{black}{\textbf{4.22GB}} & \textcolor{black}{39.58} \\
\textcolor{black}{UnifyGeo ($k=5$, default)} & \textbf{57.7M} & 38.90ms & 4.32GB & \textbf{39.64} \\
\bottomrule
\end{tabular}
}
\end{table}

\begin{table*}[t] % 表3
\centering
\caption{
Comparison with state-of-the-art methods for image retrieval on the VIGOR dataset. Under each setting, metrics are ordered as $R@1$, $R@5$, $R@10$, $R@1\%$, and Hit Rate. The best results are shown in \textbf{boldface} and the second best results are \underline{underlined}.
}
\label{tab2_retrieval_vigor}
\begin{tabular}{l|ccccc|ccccc} 
\hline\hline
\multirow{2}{*}{Methods} & \multicolumn{5}{c|}{Same-area}                                                                                                                                               & \multicolumn{5}{c}{Cross-area}                                                                                                                                               \\
                          & R@1                              & R@5                              & R@10                             & R@1\%                            & Hit Rate                         & R@1                              & R@5                              & R@10                             & R@1\%                            & Hit Rate                          \\ 
\hline
SAFA \cite{shi2019spatial} & 33.93                            & 58.42                            & 68.12                            & 98.24                            & 36.87                            & 8.20                             & 19.59                            & 26.36                            & 77.61                            & 8.85                              \\
VIGOR \textcolor{black}{\cite{zhu2021vigor}} & 41.07                            & 65.81                            & 74.05                            & 98.37                            & 44.71                            & 11.00                            & 23.56                            & 30.76                            & 80.22                            & 11.64                             \\
TransGeo \cite{zhu2022transgeo} & 61.48                            & 87.54                            & 91.88                            & \underline{{99.56}} & 73.09                            & 18.99                            & 38.24                            & 46.91                            & 88.94                            & 21.21                             \\
PaSS-KD \cite{li2023patch} & 52.90                            & 76.60                            & -                                & -                                & 57.00                            & 21.00                            & 39.70                            & -                                & -                                & 22.20                             \\
GeoDTR  \cite{zhang2023cross} & 56.51                            & 80.37                            & 86.21                            & 99.25                            & 61.76                            & 30.02                            & 52.67                            & 61.45                            & 94.40                            & 30.19                             \\
Sample4Geo \cite{deuser2023sample4geo}  & \underline{{77.86}} & \textbf{{95.66}}  & \textbf{{97.21}}  & \textbf{{99.61}}  & \textbf{{89.82}}  & \underline{{61.70}} & \textbf{{83.50}}  & \textbf{{88.00}}  & \textbf{{98.17}}  & \underline{69.87}   \\
\textcolor{black}{UnifyGeo (ours)}            & \textbf{{82.80}}  & \underline{{94.92}} & \underline{{96.57}} & 99.47                            & \underline{\textcolor{black}{89.12}} & \textbf{{67.58}}  & \underline{{81.60}} & \underline{{86.02}} & \underline{{97.26}} & \textbf{\textcolor{black}{72.26}}  \\
\hline\hline
\end{tabular}
\end{table*}

\begin{table}[t]
\centering
\caption{Comparison with state-of-the-art methods for metric localization on the VIGOR dataset.}
\label{tab3_localization_vigor}
\begin{tabular}{l|cc|cc} 
\hline\hline
\multirow{2}{*}{Methods} & \multicolumn{2}{c|}{Same-area} & \multicolumn{2}{c}{Cross-area}  \\
                          & Mean(m)       & Median(m)      & Mean(m)       & Median(m)       \\ 
\hline
VIGOR \cite{zhu2021vigor}   & 8.99          & 7.81           & 8.89          & 7.73            \\
MCC \cite{xia2022visual}  & 6.94          & 3.64           & 9.05          & 5.14            \\
SliceMatch \cite{lentsch2023slicematch}  & 5.18          & 2.58           & 5.53          & 2.55            \\
GGCVT \cite{shi2023boosting}             & 4.12          & 1.34           & 5.16          & 1.40            \\
CCVPE \cite{xia2023convolutional}        & 3.60          & 1.36           & 4.97          & 1.68            \\
HC-Net \cite{wang2024fine}& 2.65          & 1.17           & 3.36          & 1.59            \\
UnifyGeo(ours)            & \textbf{2.26} & \textbf{1.13}  & \textbf{3.05} & \textbf{1.39}   \\
\hline\hline
\end{tabular}
\end{table}

\subsection{Comparison with State-of-the-art Methods}

\textit{1) LF-CVGL Experiments on VIGOR:} 
We conduct a comprehensive evaluation of our method on the LF-CVGL task using the VIGOR dataset, considering both same-area and cross-area settings, as presented in Table \ref{tab1_loc_vigor}. 
\textcolor{black}{Our proposed UnifyGeo is first compared with two existing LF-CVGL methods, VIGOR \cite{zhu2021vigor} and PaSS-KD \cite{li2023patch}. 
To ensure a more rigorous evaluation, we further introduce two stronger baselines that combine Sample4Geo \cite{deuser2023sample4geo}, a state-of-the-art (SOTA) retrieval method, with SOTA metric localization methods, namely ``Sample4Geo+CCVPE" and ``Sample4Geo+HC-Net". 
These combined baselines leverage Sample4Geo to retrieve the top-1 satellite image and then utilize CCVPE \cite{xia2023convolutional} or HC-Net \cite{wang2024fine} to predict the query location, forming a hierarchical geo-localization pipeline akin to our proposed method. 
In addition, we report Sample4Geo under the LF-CVGL evaluation protocol as a retrieval-only reference, indicating the lower-bound performance achievable without metric localization.}

As shown in Table \ref{tab1_loc_vigor}, our method \textcolor{black}{achieves the best performance among all compared entries}. 
\textcolor{black}{Compared with PaSS-KD, a recent state-of-the-art LF-CVGL method, UnifyGeo achieves 1.75× and 4× its R@10m in the same-area and cross-area settings, respectively.}
Furthermore, at a stringent geo-location precision threshold of 1m, our method demonstrates a localization recall rate exceeding 10-fold compared to PaSS-KD in both same-area and cross-area settings. 
These substantial improvements underscore the superiority of our proposed UnifyGeo.
In addition, the retrieval-only Sample4Geo reference obtains limited meter-level localization accuracy. After being combined with CCVPE or HC-Net, its LF-CVGL performance improves substantially, indicating that retrieval alone mainly provides a coarse location estimate, while a dedicated metric localization stage is critical for high-precision LF-CVGL.

Notably, even with the improved performance achieved by the combined baselines, our method consistently yields superior results. 
Specifically, in terms of coarse precision, our method yields relative improvements of 8.6\% ($74.91\% \rightarrow 81.37\%$) and 11.8\% ($59.23\% \rightarrow 66.24\%$) over the best baseline, Sample4Geo+HC-Net, in the same-area and cross-area settings, respectively. 
For the high-precision threshold, our method achieves relative improvements of 10.4\% ($35.92\% \rightarrow 39.64\%$) in the same-area setting and 32.7\% ($19.28\% \rightarrow 25.58\%$) in the cross-area setting, respectively. 
A comprehensive comparison of LF-CVGL performance is provided in Fig. \ref{fig_4: loc_experiment}.
In addition to achieving superior localization performance, we further assess the computational efficiency of UnifyGeo against the combined baseline. 
Table~\ref{tab:efficiency_analysis} compares model parameters, per-query inference time, inference memory, and high-precision LF-CVGL localization accuracy on the VIGOR same-area setting, using an 11th Gen Intel(R) Core(TM) i7-11700KF CPU and an NVIDIA RTX 4090 GPU. 
\textcolor{black}{Compared with the cascaded baseline Sample4Geo+HC-Net, both UnifyGeo
configurations achieve better high-precision localization with fewer parameters
and lower inference memory, albeit at higher per-query latency.
The $k=3$ configuration provides lower inference latency with comparable
localization accuracy, whereas the default $k=5$ configuration achieves a
marginally higher R@1m.
The additional inference cost mainly comes from re-ranking, which exploits
fine-grained matching cues to improve aerial-reference selection and LF-CVGL
performance.
Further analysis of the efficiency and effectiveness of re-ranking is provided
in Section~IV.F.}

\begin{figure}[t]
\centering
\includegraphics[width=\linewidth]{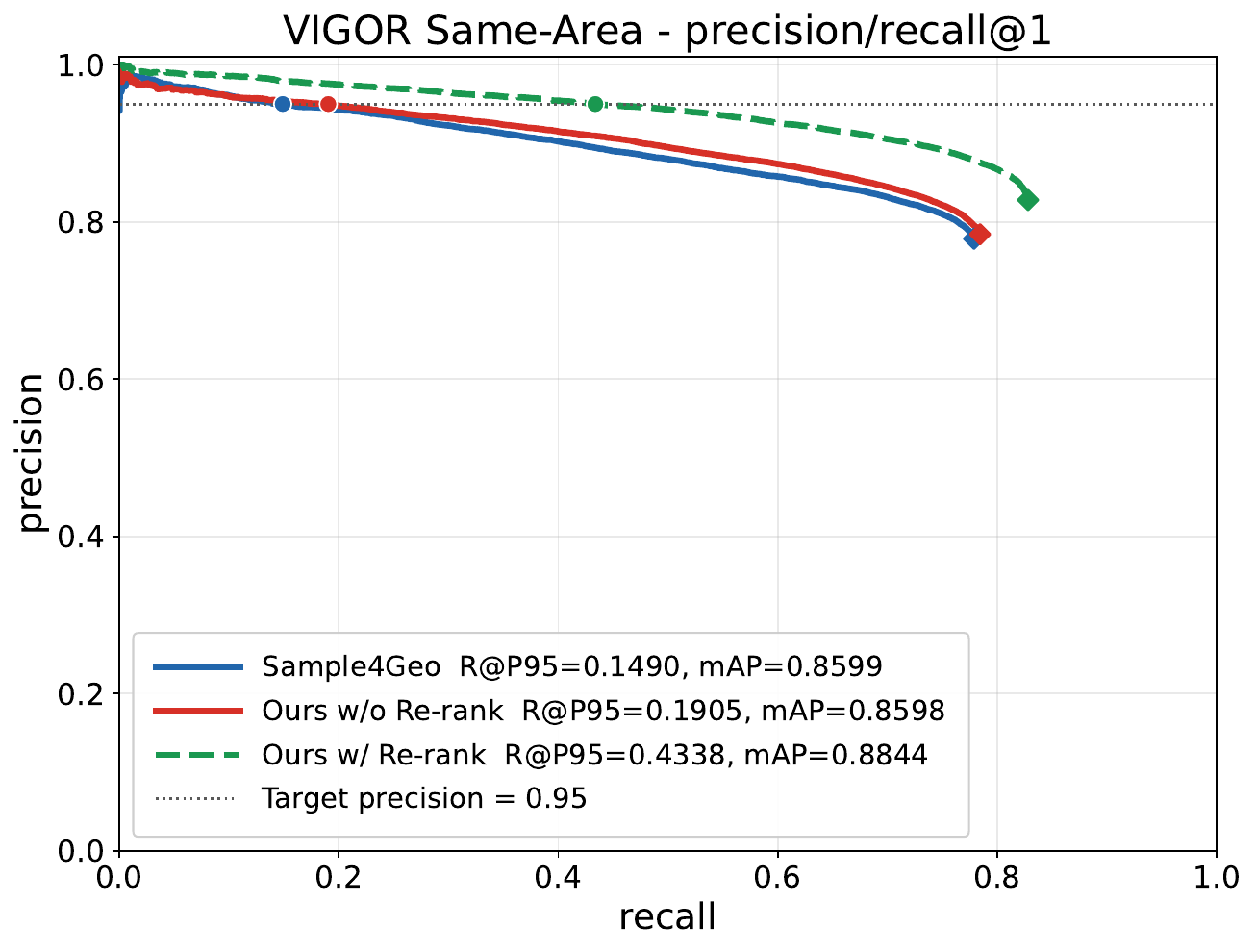}
\caption{
\textcolor{black}{Comparison of precision-recall curves for top-1 retrieval on the VIGOR same-area setting for Sample4Geo and UnifyGeo before and after re-ranking. R@P95 denotes the maximum recall achieved while maintaining at least 95\% precision, and mAP denotes retrieval mean average precision.}}
\label{fig:pr_curve}
\end{figure}

\begin{figure*}[!t]
\centering
\includegraphics[width=0.95\textwidth]{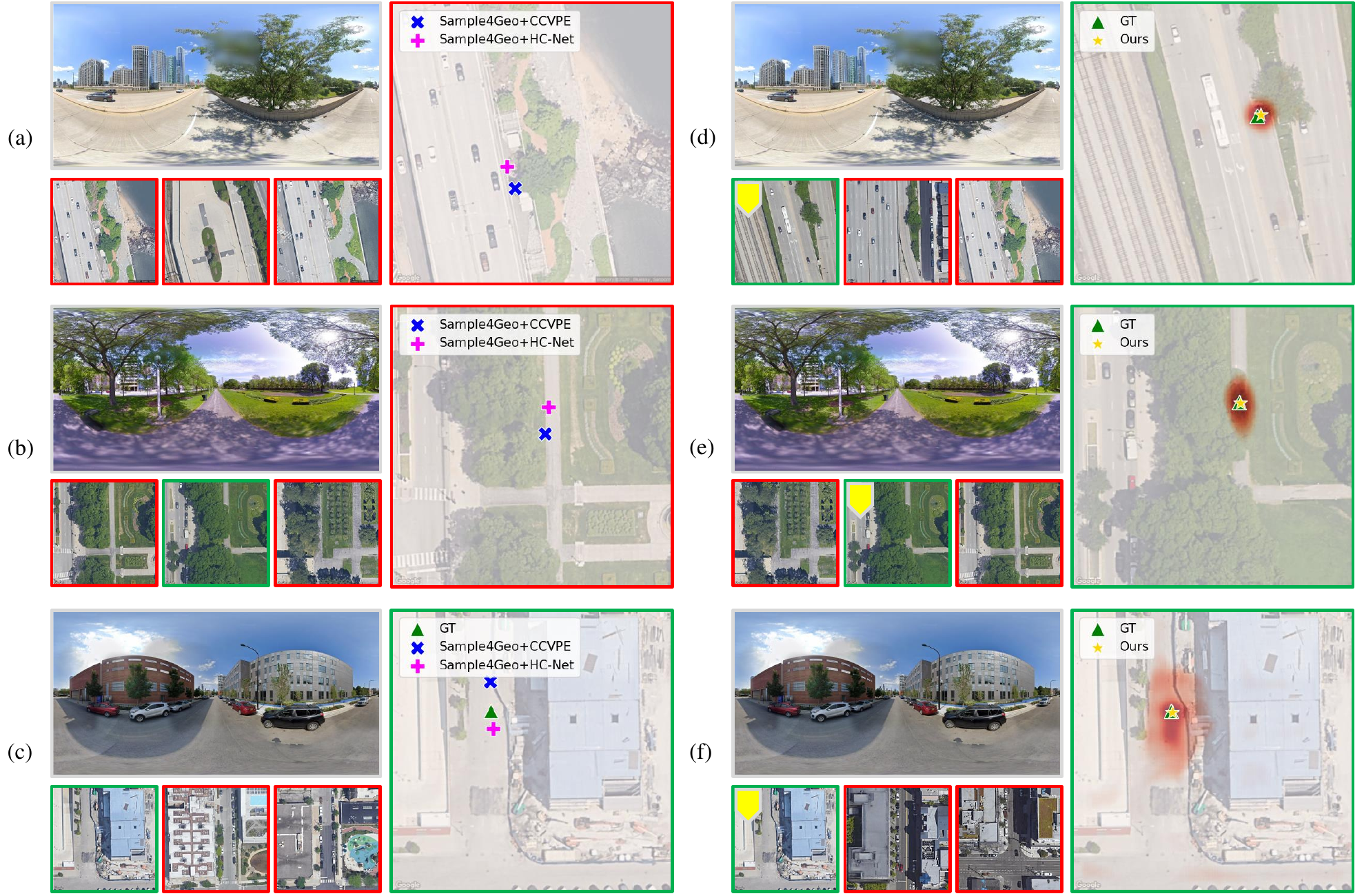}
\caption{
Visualization of geo-localization methods on VIGOR. 
The left column (a-c) presents the results of two combined baselines, while the right column (d-f) shows the results of our method. 
Each row corresponds to the geo-localization results for the same query-view. 
Each panel of (a-f) comprises: a ground-view query image (top-left), the top-3 retrieved images (bottom-left, ordered from left to right), and a localization result based on the top-1 (or re-ranked) retrieved image (right, enlarged for clarity).
Retrieved images are annotated with \textcolor{Green3}{\textbf{light green}} or \textcolor{red}{\textbf{red}} borders to indicate matches or non-matches with the query-view.
In (d-f), a ``\textbf{yellow pentagon}" on the top-left indicates the re-ranked candidate.
Localization results are marked with distinct symbols: ``\textbf{dark green triangle}" (ground-truth), ``\textbf{blue cross}" (Sample4Geo+CCVPE), ``\textbf{magenta plus}" (Sample4Geo+HC-Net), and ``\textbf{golden star}" (our method).
}
\label{fig_5: qualitative result}
\end{figure*}

\begin{table*}[t] % 表5
\centering
\caption{\textcolor{black}{Comparison} with state-of-the-art methods for image retrieval on the CVUSA and CVACT datasets. $^{\dagger}$ represents that polar transformation is applied to aerial images. 
The best results are shown in \textbf{boldface} and the second best results are \underline{underlined}.
}
\label{tab4_retrieval_cvusa}
\begin{tabular}{l|cccc|cccc|cccc} 
\hline\hline
\multirow{2}{*}{Methods} & \multicolumn{4}{c|}{CVUSA}                                                                 & \multicolumn{4}{c|}{CVACT-Val}                                                             & \multicolumn{4}{c}{CVACT-Test}                                                             \\
                          & R@1                  & R@5                  & R@10                 & R@1\%                 & R@1                  & R@5                  & R@10                 & R@1\%                 & R@1                  & R@5                  & R@10                 & R@1\%                 \\ 
\hline
SAFA$^{\dagger}$ \cite{shi2019spatial} & 89.84                & 96.93                & 98.14                & 99.64                 & 81.03                & 92.80                & 94.84                & 98.17                     & 55.50                    & 79.94                    & 85.08                    & 94.49                     \\
SEH$^{\dagger}$ \cite{guo2022soft} & 95.11                & 98.45                & 99.00                & 99.78                 & 84.75                & 93.97                & 95.46                & 98.11                     & -                    & -                    & -                    & -                     \\
TransGeo \cite{zhu2022transgeo} & 94.08                & 98.36                & 99.04                & 99.77                 & 84.95                & 94.14                & 95.78                & 98.37                 & -                    & -                    & -                    & -                     \\
PaSS-KD \cite{li2023patch} & 94.09                & 98.42                & 99.13                & 99.77                 & 87.20                & 94.30                & 95.67                & 97.85                 & 66.81                & 88.03                & 90.87                & 98.02                 \\
GeoDTR$^{\dagger}$ \cite{zhang2023cross} & 95.43                & 98.86                & 99.34                & 99.86                 & 86.21                & 95.44                & 96.72                & \textbf{{98.77}}                 & 64.52                & 88.59                & 91.96                & \textbf{{98.74}}                 \\
Sample4Geo   \cite{deuser2023sample4geo} & \underline{{98.68}} & \textbf{{99.68}}  & \textbf{{99.78}}  & \textbf{{99.87}}  & \underline{{90.81}} & \textbf{{96.74}}  & \textbf{{97.48}}  & \textbf{{98.77}}  & \underline{{71.51}} & \underline{{92.42}} & \textbf{{94.45}}  & \underline{{98.70}}  \\
\textcolor{black}{UnifyGeo (ours)}            & \textbf{{98.91}}  & \underline{{99.62}} & \underline{{99.72}} & \underline{{99.84}} & \textbf{{91.38}}  & \underline{{96.27}} & \underline{{97.04}} & \underline{{98.45}} & \textbf{{73.42}}  & \textbf{{92.46}}  & \underline{{94.27}} & 98.43                             \\
\hline\hline
\end{tabular}
\end{table*}

\textit{2) Retrieval Experiments on VIGOR:} 
To quantitatively assess the retrieval performance of our geo-localization pipeline, we present a comparative analysis of our method against SOTA methods on the VIGOR dataset, as shown in Table \ref{tab2_retrieval_vigor}. 
Our method is competitive with Sample4Geo \cite{deuser2023sample4geo} and significantly outperforms other approaches. 
Notably, our UnifyGeo achieves substantial improvements over Sample4Geo at top-1 recall, with relative improvements of 6.3\% ($77.86\% \rightarrow 82.80\%$) and 9.5\% ($61.70\% \rightarrow 67.58\%$) in the same-area and cross-area settings, respectively.
These improvements can be attributed to our effective re-ranking mechanism. 
For larger top-$k$ retrieval metrics \textcolor{black}{and same-area Hit Rate, UnifyGeo remains slightly lower than Sample4Geo, while achieving a higher cross-area Hit Rate}. These remaining gaps are largely attributable to the stronger
ConvNeXt-Base encoder adopted by Sample4Geo, whereas our default model uses the lighter ConvNeXt-Tiny for parameter efficiency. 
A detailed backbone-scale analysis, including the results of UnifyGeo with ConvNeXt-Base, is provided in Section~\uppercase\expandafter{\romannumeral4}.G.

\textcolor{black}{Following the confidence-threshold evaluation used in large-scale location recognition~\cite{chen2011city,Sattler2016CVPR}, in which
top-1 predictions are accepted according to their confidence scores, we further assess top-1 retrieval reliability on VIGOR same-area in Fig.~\ref{fig:pr_curve}. Before re-ranking, UnifyGeo improves Recall@95\% Precision from 14.90\% to 19.05\% over Sample4Geo with comparable retrieval mAP (0.8598 vs. 0.8599). Top-5 re-ranking further increases Recall@95\% Precision to 43.38\% and mAP to 0.8844. These gains demonstrate that re-ranking effectively exploits fine-grained matching cues to improve both top-1 retrieval reliability and retrieval ranking quality.}

\begingroup
\setlength{\parskip}{0pt}
\textit{3) Metric Localization Experiments on VIGOR:} 
To demonstrate the effectiveness of our metric localization, we compare the localization accuracy of our UnifyGeo with state-of-the-art methods in Table \ref{tab3_localization_vigor}. 
As shown, our method surpasses all these methods across all metrics, demonstrating exceptional performance in the metric localization task. 
\textcolor{black}{Notably, compared with HC-Net \cite{wang2024fine}, the second-best performer, UnifyGeo reduces the mean and median localization errors by $14.7\%$ and $3.4\%$, respectively, in the same-area setting, and by $9.2\%$ and $12.6\%$, respectively, in the cross-area setting.
These consistent improvements highlight the effectiveness of our model design, which enables UnifyGeo to extract more informative detailed features for metric localization.
For successfully retrieved queries, the resulting improvement in location estimation further enhances the overall LF-CVGL performance.}
\par
\endgroup

\textit{4) LF-CVGL and \sz{Retrieval} Experiments on CVUSA and CVACT:} 
To further evaluate our approach, we assess both the LF-CVGL and \sz{retrieval} performance of our method on the CVUSA and CVACT datasets, and compare it with SOTA methods in Table \ref{tab4_retrieval_cvusa}. 
Given the center-aligned nature of the data, the top-1 recall rate serves as an effective metric for LF-CVGL. 
As shown in Table \ref{tab4_retrieval_cvusa}, our UnifyGeo achieves the best geo-localization performance. 
Notably, on the CVACT-test dataset, our method yields a relative improvement of 2.7\% ($71.51\% \rightarrow 73.42\%$), demonstrating its superior generalization capabilities.
In terms of retrieval performance, our method slightly underperforms on other metrics compared to Sample4Geo, consistent with the observations on the VIGOR dataset (Table \ref{tab2_retrieval_vigor}). 
We attribute this disparity to the same underlying factors. However, our method shows a significant advantage in $R@1$, which is crucial for applications like real-time geo-localization and navigation, where the accuracy of the top-ranked result is paramount.

\begin{figure*}[!t]
\centering
\includegraphics[width=0.95\textwidth]{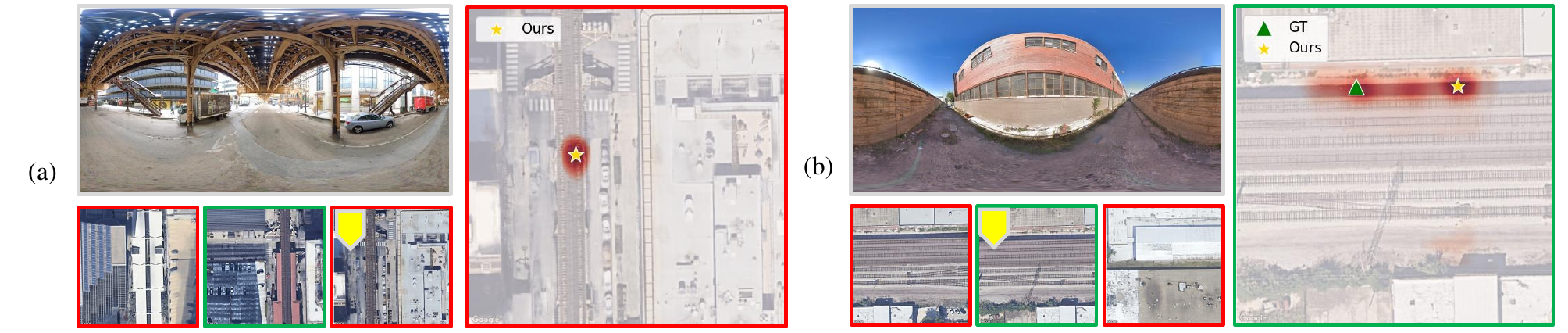}
\caption{\textcolor{black}{Failure cases of our method on VIGOR, with the same visualization layout as the qualitative examples in Fig.~\ref{fig_5: qualitative result}. 
Case (a) shows a re-ranking failure under occlusion, and Case (b) shows a localization failure in a repetitive passage area.}}
\label{fig: failure case}
\end{figure*}
\begin{table*}[t]
\centering
\caption{Ablation study for image retrieval on the VIGOR dataset. 
}
\label{tab5_ablation_retrieval}
\begin{tabular}{c|cccc|ccccc|ccccc} 
\hline\hline
\multirow{2}{*}{No.}  & \multicolumn{4}{c|}{Configurations}   & \multicolumn{5}{c|}{Same-area}                            & \multicolumn{5}{c}{Cross-area} \\ 
                     & UL                         & FA                        & Re-rank            & $\mathcal{L}_R$                    & R@1            & R@5            & R@10           & R@1\%                   & Hit Rate       & R@1            & R@5            & R@10           & R@1\%                   & Hit Rate        \\ 
\hline
1.                   & - & - & - & - & 74.12          & 93.44          & 95.62          & 99.48                   & 85.43          & 51.73          & 74.63          & 80.74          & 96.80                   & 58.10           \\
2.                   & \ding{51} & - & - & - & 75.54          & 94.08          & 96.15          & \textbf{\textbf{99.51}} & 86.53          & 53.99          & 76.31          & 82.26          & 96.98                   & 60.57           \\
3.                   & - & \ding{51} & - & - & 77.19          & 94.31          & 96.20          & 99.45                   & 87.68          & 58.83          & 80.10          & 85.01          & 97.28                   & 65.90           \\
4.                   & \ding{51} & \ding{51} & - & - & 78.01          & 94.80          & 96.48          & 99.49                   & 88.48          & 60.21          & 81.13          & 85.69          & \textbf{97.38}          & 67.53           \\
5.                   & \ding{51} & \ding{51} & - & \ding{51} & 78.44          & \textbf{94.92}          & \textbf{96.57}          & 99.47                   & 88.72        & 61.18          & \textbf{81.60}          & \textbf{86.02}          & 97.26          & 68.24           \\
6.                   & \ding{51} & \ding{51} & \ding{51} & - & 80.97          & 94.80          & 96.48          & 99.49                   & \textcolor{black}{88.51}          & 64.57          & 81.13          & 85.69          & \textbf{\textbf{97.38}} & \textcolor{black}{69.35}           \\
Ours                 & \ding{51} & \ding{51} & \ding{51} & \ding{51} & \textbf{82.80} & \textbf{94.92} & \textbf{96.57} & 99.47                   & \textcolor{black}{\textbf{89.12}} & \textbf{67.58} & \textbf{81.60} & \textbf{86.02} & 97.26                   & \textbf{\textcolor{black}{72.26}}  \\
\hline\hline
\end{tabular}
\end{table*}

\subsection{Qualitative Analysis}

To provide a more intuitive demonstration of our method's effectiveness, we visualize the LF-CVGL results of our method and two combined baselines, ``Sample4Geo+CCVPE" and ``Sample4Geo+HC-Net", on the VIGOR dataset in Fig. \ref{fig_5: qualitative result}. The comparison highlights the superiority of our method in three distinct scenarios. Specifically, as shown in Fig. \ref{fig_5: qualitative result}(a) and \ref{fig_5: qualitative result}(d), our method successfully localizes the image, whereas the combined baselines fail to retrieve the correct reference. In Fig. \ref{fig_5: qualitative result}(b) and \ref{fig_5: qualitative result}(e), despite incorrect top-1 image initially retrieved by all methods, our method's re-ranking mechanism enables correct selection of the satellite image and successful localization. Furthermore, in Fig. \ref{fig_5: qualitative result}(c) and \ref{fig_5: qualitative result}(f), even when all methods retrieve the correct image, our method outperforms the combined baselines in terms of localization accuracy, demonstrating its enhanced precision and reliability, and ultimately leading to more accurate geo-localization.

\textcolor{black}{We further show failure cases of our method in Fig.~\ref{fig: failure case}. 
In Fig.~\ref{fig: failure case}(a), occlusion in the upper region of the ground-view image limits visible scene information, causing re-ranking to select an incorrect aerial reference and leading to localization on the wrong image. 
In Fig.~\ref{fig: failure case}(b), although re-ranking selects the correct aerial reference, the query lies in a long passage area with repetitive visual patterns, where severe spatial ambiguity leads to a large localization shift. 
These cases indicate that UnifyGeo may still be challenged by limited ground-view observations and highly repetitive structures.}

\subsection{Ablation Studies}

In this section, we demonstrate the effectiveness of our proposed components on VIGOR, including the Unified Learning Strategy (UL), Feature Aggregator (FA), Re-ranking process, and the re-ranking loss $\mathcal{L}_R$.
To ensure a thorough analysis, we conduct ablation studies on all three tasks: image retrieval, metric localization, and LF-CVGL. The results of these evaluations are presented in Tables \ref{tab5_ablation_retrieval}, \ref{tab6_ablation_loc}, and \ref{tab7_ablation_loc_all}, respectively.
\textcolor{black}{For Tables \ref{tab5_ablation_retrieval} and \ref{tab6_ablation_loc}, we construct the task-specific baselines by independently training the global descriptor extraction pathway with average pooling and the detailed feature extraction pathway together with the localization decoder, respectively. For Table VIII, these two independently trained models are cascaded to form the LF-CVGL baseline.}
The corresponding baseline results are reported in the first row of each table.
After analyzing the effects of individual components, we further investigate the candidate number $k$ to evaluate the efficiency and effectiveness of re-ranking.

\begin{table}[t]
\centering
\caption{Ablation study for metric localization on the VIGOR dataset.
}
\label{tab6_ablation_loc}
\resizebox{\columnwidth}{!}{%
\begin{tabular}{c|ccc|cc|cc} 
\hline\hline
\multirow{2}{*}{No.} & \multicolumn{3}{c|}{Configurations}                                                                                         & \multicolumn{2}{c|}{Same-area} & \multicolumn{2}{c}{Cross-area}  \\
                     & UL                                                   & FA          & $\mathcal{L}_{R}$                                     & Mean(m)       & Median(m)      & Mean(m)       & Median(m)       \\ 
\hline
1.                   & -                                           & - & -                                           & 2.59          & \textbf{1.00}  & 4.30          & 1.50            \\
2.                   & \ding{51}                                           & - & -                                           & 2.63          & 1.18           & 4.22          & 1.43            \\
3.                   & \ding{51}                                           & \ding{51} & \begin{tabular}[c]{@{}c@{}}-\\\end{tabular} & 2.41          & 1.12           & 3.31          & 1.42            \\
Ours                 & \begin{tabular}[c]{@{}c@{}}\ding{51}\\\end{tabular} & \ding{51} & \ding{51}                                           & \textbf{2.26} & 1.13           & \textbf{3.05} & \textbf{1.39}   \\
\hline\hline
\end{tabular}
}
\end{table}

\begin{table}[t]
\centering
\caption{Ablation study for LF-CVGL on the VIGOR dataset.
\textcolor{black}{No.~1 cascades our separately trained retrieval and standalone localization
models from Table~VI No.~1 and Table~VII No.~1, respectively.
``Params.'' is the number of model parameters.}
}
\label{tab7_ablation_loc_all}
\resizebox{\columnwidth}{!}{%
\begin{tabular}{c|cccc|c|cc|cc}
\hline\hline
\multirow{2}{*}{No.}
& \multicolumn{4}{c|}{Configurations}
& \multirow{2}{*}{\textcolor{black}{Params.}}
& \multicolumn{2}{c|}{Same-area}
& \multicolumn{2}{c}{Cross-area} \\
& UL
& FA
& Re-rank
& $\mathcal{L}_{R}$
&
& R@1m
& R@10m
& R@1m
& R@10m \\
\hline
1.
& -
& -
& -
& -
& \textcolor{black}{113.0M}
& \textcolor{black}{39.33}
& \textcolor{black}{71.19}
& \textcolor{black}{17.96}
& \textcolor{black}{49.05}\\

2.
& \ding{51}
& -
& -
& -
& \textcolor{black}{55.3M}
& 34.64
& 73.51
& 19.48
& 52.24 \\

3.
& \ding{51}
& \ding{51}
& -
& -
& \textcolor{black}{57.7M}
& 36.63
& 76.07
& 21.88
& 58.46 \\

4.
& \ding{51}
& \ding{51}
& -
& \ding{51}
& \textcolor{black}{57.7M}
& 37.37
& 76.74
& 22.81
& 59.63 \\

5.
& \ding{51}
& \ding{51}
& \ding{51}
& -
& \textcolor{black}{57.7M}
& 38.48
& 79.47
& 23.98
& 63.17 \\

Ours
& \ding{51}
& \ding{51}
& \ding{51}
& \ding{51}
& \textcolor{black}{57.7M}
& \textbf{39.64}
& \textbf{81.37}
& \textbf{25.58}
& \textbf{66.24} \\
\hline\hline
\end{tabular}%
}
\end{table}

\begin{table*}[t]
\centering
\caption{Ablation study on the candidate number $k$ for re-ranking on the VIGOR same-area setting. ``N/A'' denotes no effective re-ranking, and ``--'' indicates that the corresponding $R@K$ metric is unaffected by re-ranking among top-$k$ candidates.}
\label{tab:rerank_efficiency}
\setlength{\tabcolsep}{1pt}
\begin{tabular}{cccccccccc}
\hline\hline
\multirow{2}{*}{$k$} 
& \multirow{2}{*}{Retrieval Time} 
& \multirow{2}{*}{Re-rank Time} 
& \multirow{2}{*}{Loc. Time} 
& \multicolumn{3}{c}{Retrieval} 
& \multicolumn{2}{c}{LF-CVGL} \\
\cmidrule(lr){5-7}
\cmidrule(lr){8-9}
& & & & $R@1$ & $R@5$ & $R@10$ & $R@1m$ & $R@10m$ \\
\midrule
1 & 7.0ms & N/A & 6.5ms & 78.44 & 94.92 & 96.57 & 37.37 & 76.74 \\
3 & 7.0ms & 15.7ms & 6.5ms & \textbf{82.91} & -- & -- & 39.58 & \textbf{81.40} \\
5 (\textcolor{black}{default}) & 7.0ms & 25.4ms & 6.5ms & 82.80 & -- & -- & \textbf{39.64} & 81.37 \\
10 & 7.0ms & 48.2ms & 6.5ms & 82.58 & \textbf{95.23} & -- & 39.62 & 81.21 \\
20 & 7.0ms & 99.9ms & 6.5ms & 82.39 & 94.85 & \textbf{96.60} & 39.61 & 81.05 \\
\hline\hline
\end{tabular}
\end{table*}

\textit{1) Effects of UL:} 
As shown in Table \ref{tab5_ablation_retrieval} (rows No. 1 vs. 2), joint training with detailed features improves retrieval performance across all metrics compared to the baseline that only trains global descriptor extraction pathway.
A similar trend is also observed in Nos. 3 vs. 4.
However, UL alone does not guarantee consistent improvements in metric localization.
As shown in Table~\ref{tab6_ablation_loc} (Nos. 1 vs. 2),
\textcolor{black}{UL reduces the localization errors in the cross-area setting but leads to degraded localization performance in the same-area setting.
This trade-off is also reflected in the complete LF-CVGL pipeline. As shown in Table~\ref{tab7_ablation_loc_all} (Nos. 1 vs. 2), compared with the cascade of independently trained models, UL improves R@10m in both settings and R@1m in the cross-area setting, but decreases R@1m in the same-area setting. 
The contrasting performance trends observed in the same-area and cross-area settings suggest a potential optimization conflict between global retrieval learning and fine-grained spatial matching within the shared encoder, which motivates the introduction of FA.
Despite this trade-off, UL integrates retrieval and metric localization into a single network, substantially reducing the parameter count and avoiding the overhead of separately training and deploying two task-specific models.
Overall, these results demonstrate the effectiveness of UL in enabling a compact and unified model for LF-CVGL.}

\begin{figure}[t]
\centering
\includegraphics[width=0.8\linewidth]{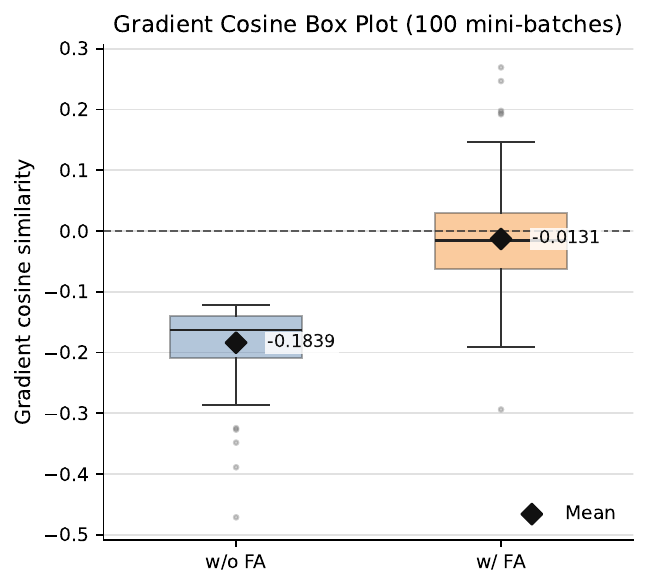}
\caption{
\textcolor{black}{Gradient-cosine distribution on conflict-intensive mini-batches. We evaluate 1,000 mini-batches from the VIGOR training split and visualize the 100 mini-batches with the most negative no-FA gradient cosine. The cosine similarity is computed between the retrieval and metric localization gradients with respect to the shared aerial encoder. The black diamond denotes the mean, the center line denotes the median, and the box denotes the interquartile range.}
}
\label{fig:grad_cos}
\end{figure}

\textit{2) Effects of FA:} 
As shown in Table~\ref{tab5_ablation_retrieval}, replacing average pooling with our proposed FA module generally improves retrieval performance (Nos. 1 vs. 3 and Nos. 2 vs. 4).
\textcolor{black}{For metric localization, adding FA to the UL-only configuration reduces both the mean and median errors in the same-area and cross-area settings (Nos. 2 vs. 3 in Table~\ref{tab6_ablation_loc}), indicating improved compatibility between retrieval and metric localization.}
To directly quantify \textcolor{black}{this effect}, we compute the gradient cosine similarity between the two objectives with respect to the shared aerial encoder.
A negative cosine indicates conflicting update directions, while a positive cosine suggests aligned optimization.
\textcolor{black}{As shown in Fig.~\ref{fig:grad_cos}, for the mini-batches exhibiting the strongest conflict without FA, introducing FA shifts the gradient-cosine distribution substantially toward zero and positive alignment, with 42 of the 100 mini-batches obtaining positive cosine values.}
\textcolor{black}{These results demonstrate that FA substantially alleviates the conflict and enables aligned optimization for a considerable portion of samples. The benefits of FA also extend to the complete LF-CVGL pipeline. As shown in Table~\ref{tab7_ablation_loc_all} (Nos. 2 vs. 3), FA improves all four LF-CVGL metrics with only a modest increase in the parameter count.}
\textcolor{black}{Nevertheless, FA does not completely eliminate the trade-off introduced by unified learning. In Table~\ref{tab6_ablation_loc}, the same-area median error increases from 1.00 m for the standalone localization baseline to 1.18 m with UL, is partially recovered to 1.12 m after applying FA, and reaches 1.13 m in the full model. Consistently, the UL+FA configuration in Table~\ref{tab7_ablation_loc_all} achieves a lower same-area R@1m than the independently trained cascade, although it achieves higher same-area R@10m and better cross-area performance.
Applying FA leads to larger reductions in mean errors than in median errors, suggesting that it mainly reduces large errors on challenging samples, whereas the standalone localization model retains better typical-case precision in the same-area setting.
In the complete LF-CVGL pipeline, however, the cascaded baseline relies on separately trained retrieval and localization models, resulting in a substantially higher parameter count and additional training and deployment overhead. The FA-equipped unified model instead remains compact and shows greater robustness to city-specific appearance variations, as reflected by the marked reduction in cross-area mean error.
}
\vspace{-1pt}
\textit{3) Effects of Re-ranking:} 
In our LF-CVGL pipeline, re-ranking plays a crucial role in improving $R@1$ accuracy and informing metric localization. 
As shown in Table \ref{tab5_ablation_retrieval} (Nos. 4 vs. 6), re-ranking yields a relative improvement of $3.8\%$ and $7.2\%$ at $R@1$ metric in the same-area and cross-area settings, respectively.
\textcolor{black}{Re-ranking also improves Hit Rate, particularly in the more challenging VIGOR cross-area setting. Since subsequent metric localization is performed on the top-ranked aerial image, the improvement in R@1 increases the probability of providing the correct aerial reference for localization. Consequently, re-ranking improves LF-CVGL performance,} yielding relative improvements of $5.1\%$ and $9.6\%$ in $R@1m$, as well as $4.5\%$ and $8.1\%$ in $R@10m$ metrics under the same-area and cross-area settings, respectively, as shown in Table \ref{tab7_ablation_loc_all} (\textcolor{black}{Nos. 3 vs. 5}).
These results highlight the effectiveness of re-ranking, which relies on detailed feature matching and further underscore the importance of our unified multi-granularity feature learning strategy.

\textit{4) Effects of $\mathcal{L}_R$ Loss:} 
The last two rows in Tables \ref{tab5_ablation_retrieval} and \ref{tab6_ablation_loc} present the performance of our UnifyGeo framework before and after applying the $\mathcal{L}_R$ loss.
The results indicate that incorporating $\mathcal{L}_R$ not only improves retrieval performance (with re-ranking) but also enhances metric localization accuracy.
The improvement on the metric localization task can be attributed to the expansion of the comparison scope of detailed features from the image level to the mini-batch level. 
This expansion increases the number of negative samples in the InfoNCE loss calculation, thereby enhancing feature distinctiveness and enabling more accurate localization through improved initial matching scores.
Furthermore, the use of $\mathcal{L}_R$ also \sz{boosts} retrieval performance (without re-ranking), as seen in Table \ref{tab5_ablation_retrieval} (Nos. 4 vs. 5). 
This enhancement is due to the refinement of detailed features, which strengthens the shared encoder's ability to capture cross-view key patterns and improves the discriminative power of global descriptors.
\sz{
These coordinated improvements in retrieval and metric localization further highlight the effectiveness of our unified framework.
}
Ultimately, our method, which combines all proposed components, achieves the best LF-CVGL performance on both settings across all localization metrics, as shown in Table \ref{tab7_ablation_loc_all}.

\textit{5) Effects of Candidate Number $k$:}
We further analyze the effect of the candidate number $k$ on re-ranking cost, retrieval accuracy, and LF-CVGL performance in Table~\ref{tab:rerank_efficiency}.\footnote{\textcolor{black}{Table~\ref{tab:rerank_efficiency} reports the run used in the main
experiments. Across three independently trained models with different random
seeds, the standard deviations of the accuracy metrics range from 0.02 to
0.15 percentage points over the evaluated values of $k$.}}
Since varying $k$ only affects candidate selection, standalone metric localization is not separately reported.
As shown in Table~\ref{tab:rerank_efficiency}, the case of $k=1$ corresponds to no effective re-ranking, since only one aerial candidate is retained and the candidate order cannot be changed. 
In this setting, UnifyGeo achieves comparable inference time to the cascaded baseline Sample4Geo+HC-Net in Table~\ref{tab:efficiency_analysis}, while obtaining better high-precision LF-CVGL accuracy. 
As $k$ increases from 3 to 20, the re-ranking time increases because detailed
matching needs to be performed for more aerial candidates, while the
localization performance varies only slightly.
A larger $k$ provides more candidates for correction, but also introduces more
visually similar distractors, making it more difficult for detailed matching
to identify the correct aerial reference.
Therefore, increasing $k$ does not necessarily improve \textcolor{black}{retrieval $R@1$ or} LF-CVGL performance.
\textcolor{black}{As shown in Table~\ref{tab:rerank_efficiency}, the results
with $k=3$ and $k=5$ show comparable overall accuracy while offering different
trade-offs between localization accuracy and computational efficiency.
Specifically, $k=3$ incurs a lower re-ranking cost, whereas $k=5$ achieves a
marginally higher $R@1m$. Accordingly, we retain $k=5$ as the default setting
for the main experiments and additionally report $k=3$ as an
efficiency-oriented configuration.}

\vspace{-0.3cm}
\sz{\subsection{Architectural Analysis}}

\sz{\textit{1) Effects of Multi-granularity Representations:} 
To validate the architectural advantage of UnifyGeo over global-descriptor-based LF-CVGL paradigms, we compare several variants on the VIGOR same-area setting in Table~\ref{tab:global_descriptor_loc}. 
First, we follow the localization paradigm of PaSS-KD~\cite{li2023patch} under our training setting by removing the fine-grained branch and using global descriptors with an MLP head for coordinate regression. 
Although this variant, denoted as ``Ours w/ MLP loc.'', improves retrieval R@1 over PaSS-KD from 52.90\% to 61.72\%, its LF-CVGL R@1m only reaches 9.06\%, showing that global descriptors with direct offset regression are insufficient for meter-level localization. 
Second, we replace the MLP head with our localization decoder while still using projected global descriptors as decoder inputs. 
This variant, denoted as ``Ours w/o F-head'', improves R@1m to 33.79\%, but still lags behind full UnifyGeo, especially in re-ranking R@1 and high-precision LF-CVGL accuracy. 
These results show that UnifyGeo's superiority comes not only from the localization decoder, but also from the coordinated use of multi-granularity representations, where detailed features are essential for reliable re-ranking and high-precision dense localization.}

\begin{table}[t]
\centering
\caption{
{
Analysis of multi-granularity representations on VIGOR.
``Ours w/ MLP loc.'' denotes global-descriptor-based coordinate regression, and ``Ours w/o F-head'' denotes using projected global descriptors in the localization decoder.
}
}
\label{tab:global_descriptor_loc}
\setlength{\tabcolsep}{4.2pt}
\begin{tabular}{lcccc}
\toprule
{\multirow{2}{*}{Method}} & {Retrieval} & {\textcolor{black}{Re-rank}} & \multicolumn{2}{c}{{LF-CVGL}} \\
\cmidrule(lr){2-2} \cmidrule(lr){3-3} \cmidrule(lr){4-5}
{} & {R@1} & {R@1} & {R@1m} & {R@10m} \\
\midrule
{PaSS-KD~\cite{li2023patch}} & {52.90} & {-} & {1.53} & {46.50} \\
{Ours w/ MLP loc.} & {61.72} & {-} & {9.06} & {56.57} \\
{Ours w/o F-head} & {77.27} & {81.82} & {33.79} & {80.31} \\
{Ours} & {\textbf{78.44}} & {\textbf{82.80}} & {\textbf{39.64}} & {\textbf{81.37}} \\
\bottomrule
\end{tabular}
\end{table}

\sz{\textit{2) Effects of Sharing Depth in the Feature Encoder:} 
The shared encoder is designed to construct task associations between retrieval and metric localization, but different sharing depths may lead to different trade-offs between shared feature learning and task-specific adaptation.
To validate this architectural choice, we compare different parameter sharing strategies in Table~\ref{tab:sharing_strategy}.
All variants are trained with the unified learning strategy and without the re-ranking loss to isolate the influence of feature sharing. 
We compare three configurations: sharing Stages $1$-$3$ together with both extractor heads, sharing Stages $1$-$3$, and sharing only Stages $1$-$2$. 
We also evaluate the effect of the proposed FA module under each sharing configuration.}

\textcolor{black}{Rows 1 to 3 show that, without FA, different sharing strategies lead to trade-offs between retrieval and metric localization. 
Overall, later branching improves retrieval $R@1$, whereas earlier branching after Stage $2$ slightly improves localization accuracy.
\textcolor{black}{Comparing rows 1 to 3 with rows 4 to 6, FA consistently improves retrieval performance and reduces the mean localization error, with the only exception being a slight increase from 2.49 m to 2.51 m when Stages $1$-$3$ and both extractor heads are shared. Overall, these results indicate that FA generally improves task compatibility across different sharing strategies.}
Among all variants, sharing Stages $1$-$3$ with FA achieves the best retrieval performance while maintaining competitive localization accuracy. 
Although sharing only Stages $1$-$2$ with FA slightly reduces the mean error, it increases the parameter count from $57.7$M to $79.9$M and leads to lower retrieval performance. 
Therefore, we adopt sharing Stages $1$-$3$ with task-specific heads and FA for the best overall trade-off among retrieval accuracy, localization accuracy, and parameter efficiency.}

\begin{table}[t]
\centering
\caption{Parameter sharing analysis on the VIGOR same-area setting. ``S'' denotes stage, and ``heads'' denotes the task-specific extractor heads.}
\label{tab:sharing_strategy}
\setlength{\tabcolsep}{2.1pt}
\begin{tabular}{cccccccc}
\toprule
\multirow{2}{*}{No.} 
& \multirow{2}{*}{Shared Parts} 
& \multirow{2}{*}{FA} 
& \multirow{2}{*}{Params.} 
& \multicolumn{2}{c}{Retrieval} 
& \multicolumn{2}{c}{Metric \textcolor{black}{Localization}} \\
\cmidrule(lr){5-6}
\cmidrule(lr){7-8}
& & & & $R@1$ & $R@5$ & Mean(m) & Median(m) \\
\midrule
1 & S1--S3, heads & -- & \textbf{43.4M} & 75.46 & 93.97 & 2.49 & 1.19 \\
2 & S1--S3       & -- & 55.3M & 75.54 & 94.08 & 2.63 & 1.18 \\
3 & S1--S2       & -- & 77.6M & 75.03 & 93.83 & 2.48 & \textbf{1.06} \\
4 & S1--S3, heads & $\checkmark$ & 45.8M & 77.76 & 94.68 & 2.51 & 1.15 \\
5 & S1--S3 (ours)& $\checkmark$ & 57.7M & \textbf{78.01} & \textbf{94.80} & 2.41 & 1.12 \\
6 & S1--S2       & $\checkmark$ & 79.9M & 77.49 & 94.72 & \textbf{2.40} & \textbf{1.06} \\
\bottomrule
\end{tabular}
\end{table}

\begin{table}[t]
\centering
\caption{Analysis of backbone scale on the VIGOR same-area setting. ``S4G'' denotes Sample4Geo, and ``Shared'' denotes the variant with
weight-shared retrieval branches.}
\label{tab:retrieval_arch_analysis}
\scriptsize
\setlength{\tabcolsep}{2.2pt}
\renewcommand{\arraystretch}{1.05}
\begin{tabular}{@{}lcccccccc@{}}
\toprule
\multirow{2}{*}{Method}
& \multirow{2}{*}{Params.}
& \multicolumn{4}{c}{Retrieval}
& \multicolumn{2}{c}{Metric Loc.}
& \multicolumn{1}{c}{LF-CVGL} \\
\cmidrule(lr){3-6}
\cmidrule(lr){7-8}
\cmidrule(l){9-9}
& & R@1 & R@5 & R@10 & \textcolor{black}{Hit}
& Mean & Med. & R@1m \\
\midrule
S4G (Tiny)
& 28.5M & 74.33 & 93.63 & 95.83 & \textcolor{black}{84.77}
& -- & -- & 0.23 \\

S4G (Base)
& 88.5M & 77.86 & 95.66 & 97.21 & \textcolor{black}{89.82}
& -- & -- & 0.24 \\

Ours (Shared)
& 38.2M & 83.25 & 95.11 & 96.68 & \textcolor{black}{90.98}
& 2.32 & 1.19 & 36.63 \\

Ours (Tiny)
& 57.7M & 82.80 & 94.92 & 96.57 & \textcolor{black}{89.12}
& 2.26 & 1.13 & 39.64 \\

Ours (Base)
& 180M
& \textbf{85.59}
& \textbf{95.99}
& \textbf{97.32}
& \textbf{\textcolor{black}{92.92}}
& \textbf{1.93}
& \textbf{1.02}
& \textbf{41.24} \\
\bottomrule
\end{tabular}
\end{table}

\begin{table}[t]
\centering
\caption{Sensitivity analysis of the re-ranking fusion weight. $R@1$ and $R@1m$ denote re-ranked top-1 recall and high-precision LF-CVGL accuracy, respectively.}
\label{tab:rerank_weight}
\resizebox{0.95\linewidth}{!}{
\begin{tabular}{llccccc}
\toprule
\multirow{2}{*}{Dataset} 
& \multirow{2}{*}{Metric} 
& \multicolumn{5}{c}{Fusion weight $\eta$} \\
\cmidrule(lr){3-7}
& & 0.3 & 0.4 & 0.5 & 0.6 & 0.7 \\
\midrule
\multirow{2}{*}{Same-area} 
& $R@1$  & 83.13 & \textbf{83.14} & 82.80 & 81.95 & 80.68 \\
& $R@1m$ & 39.46 & \textbf{39.67} & 39.64 & 39.43 & 39.04 \\
\midrule
\multirow{2}{*}{Cross-area} 
& $R@1$  & 67.96 & \textbf{68.03} & 67.58 & 66.62 & 65.56 \\
& $R@1m$ & 25.51 & 25.57 & \textbf{25.58} & 25.24 & 24.93 \\
\midrule
CVACT-Val 
& $R@1$  & 91.21 & 91.30 & 91.38 & 91.40 & \textbf{91.45} \\
CVUSA 
& $R@1$  & 98.85 & 98.88 & \textbf{98.91} & 98.87 & \textbf{98.91} \\
\bottomrule
\end{tabular}}
\end{table}

\textcolor{black}{\textit{3) Effects of Backbone Scale:}
To analyze how backbone scale affects UnifyGeo, we evaluate our method with different ConvNeXt backbones in Table~\ref{tab:retrieval_arch_analysis}. 
Since Table~\ref{tab2_retrieval_vigor} shows that UnifyGeo is slightly lower than Sample4Geo \textcolor{black}{in larger top-$k$ retrieval metrics and same-area Hit Rate}, we report Sample4Geo with ConvNeXt-Tiny and ConvNeXt-Base for reference. 
The results indicate that this gap is largely related to the stronger ConvNeXt-Base backbone used by Sample4Geo, while our default UnifyGeo adopts the lighter ConvNeXt-Tiny for efficiency. 
With the same backbone scale, \textcolor{black}{UnifyGeo outperforms Sample4Geo across all
retrieval metrics} (rows 1 vs. 4 and rows 2 vs. 5), and a larger backbone further improves UnifyGeo across retrieval, metric localization, and LF-CVGL.}

Moreover, Sample4Geo suggests that sharing the ground and aerial retrieval branches can improve retrieval performance. 
We therefore evaluate a shared retrieval-branch variant based on the default Tiny model to examine whether this strategy benefits the full LF-CVGL pipeline. 
As shown in row 3, this variant improves retrieval accuracy and reduces parameters, but rows 3 and 4 show that the default configuration achieves better metric localization and LF-CVGL performance, which is more important for fine-grained geo-localization. 
Therefore, we keep separate retrieval branches and adopt ConvNeXt-Tiny in the default UnifyGeo for a better accuracy-efficiency trade-off.

\textcolor{black}{\textit{4) Effects of Re-ranking Fusion Weight:}
We further analyze the influence of the fusion weight between the global retrieval score and the detailed matching score in re-ranking. 
Specifically, the weighted re-ranking score is formulated as
\begin{equation}
(1-\eta)\operatorname{sim}(v_g, v_a^t) + \eta \max_{(i,j)} \mathcal{M}_t^h(i,j),
\end{equation}
where $\eta$ controls the contribution of the detailed matching score. 
When $\eta=0.5$, the two scores are equally weighted, which is equivalent to the direct summation used in Eq.~\ref{eq:re-ranking-sum}, since multiplying the whole score by the same positive constant does not change the ranking order.}

\textcolor{black}{As shown in Table~\ref{tab:rerank_weight}, different fusion weights only lead to slight performance variations across datasets and metrics. 
The best-performing weight is not consistent across different evaluation settings, while $\eta=0.5$ achieves consistently competitive performance. 
This indicates that the proposed re-ranking formulation is stable and not sensitive to moderate changes in the fusion weight. 
Therefore, we adopt $\eta=0.5$ as a simple and balanced default setting for all experiments, which avoids dataset-specific tuning while maintaining strong re-ranking and LF-CVGL performance.}

\vspace{0.4cm}
\section{Conclusion}
This paper proposes UnifyGeo, a unified hierarchical framework that integrates image retrieval and metric localization to achieve LF-CVGL in a coarse-to-fine manner.
We design a unified learning strategy to jointly learn multi-granularity representations and propose a feature aggregator to alleviate conflicts in representation
learning \textcolor{black}{and improve the compatibility between retrieval and metric localization during joint optimization.}
In addition, we further enhance the task coordination within UnifyGeo using a re-ranking mechanism, which provides better references for metric localization from retrieval candidates.
Extensive experiments demonstrate the superiority of our proposed UnifyGeo framework on LF-CVGL, as well as its effectiveness on individual retrieval and metric localization tasks.
\sz{Future work will focus on improving robustness to challenging failure cases, such as highly repetitive structures and severe occlusion.} 
We will also explore \sz{the} adaptability \sz{of UnifyGeo} to sequential ground images and optimize the framework for lightweight deployment, aiming to develop a more universal geo-localization solution.

\section*{Acknowledgments}
This work was partially supported by the National Natural Science Foundation of China (No. 62372491), the Guangdong Basic and Applied Basic Research Foundation (2022B1515020103, 2023B1515120087), the Shenzhen Science and Technology Program (No. RCYX20200714114641140).%, and the SYSU-Sendhui Joint Lab on Embodied AI.

\bibliographystyle{IEEEtran} 
\bibliography{IEEEabrv,reference}

\end{document}